%% file: main.tex
\documentclass[10pt,twocolumn,letterpaper]{article}

\usepackage{cvpr}              

\input{preamble}

\definecolor{cvprblue}{rgb}{0.21,0.49,0.74}
\usepackage[pagebackref,breaklinks,colorlinks,citecolor=cvprblue]{hyperref}
\usepackage{graphicx}
\usepackage{amsmath}
\usepackage{amssymb}
\usepackage{booktabs}
\usepackage{adjustbox,multirow,wasysym,MnSymbol}
\def\paperID{2} 
\def\confName{CVPR}
\def\confYear{2024}

\title{Do More With What You Have: Transferring Depth-Scale from Labeled to Unlabeled Domains}

\author{Alexandra Dana\and Nadav Carmel\and Amit Shomer\and Ofer Manela\and Tomer Peleg  \\
	Samsung Israel R\&D Center \\
	Tel Aviv, Israel\\
	{\tt\small alex.dana@samsung.com}
}

\begin{document}
\maketitle

\begin{abstract}
	Transferring the absolute depth prediction capabilities of an estimator to a new domain is a task with significant real-world applications. This task is specifically challenging when images from the new domain are collected without ground-truth depth measurements, and possibly with sensors of different intrinsics. To overcome such limitations, a recent zero-shot solution was trained on an extensive training dataset and encoded the various camera intrinsics. Other solutions generated synthetic data with depth labels that matched the intrinsics of the new target data to enable depth-scale transfer between the domains. 
	
	In this work we present an alternative solution that can utilize any existing synthetic or real dataset, that has a small number of images annotated with ground truth depth labels. Specifically, we show that self-supervised depth estimators result in up-to-scale predictions that are linearly correlated to their absolute depth values across the domain, a property that we model in this work using a single scalar. In addition, aligning the field-of-view of two datasets prior to training, results in a common linear relationship for both domains. We use this observed property to transfer the depth-scale from source datasets that have absolute depth labels to new target datasets that lack these measurements, enabling absolute depth predictions in the target domain.
	
	The suggested method was successfully demonstrated on the KITTI, DDAD and nuScenes datasets, while using other existing real or synthetic source datasets, that have a different field-of-view, other image style or structural content, achieving comparable or better accuracy than other existing methods that do not use target ground-truth depths.
\end{abstract}

\vspace{-5pt}
\section{Introduction}
\label{intro}
\begin{figure}[t]
	\begin{center}
		\includegraphics[width=1.25\linewidth]{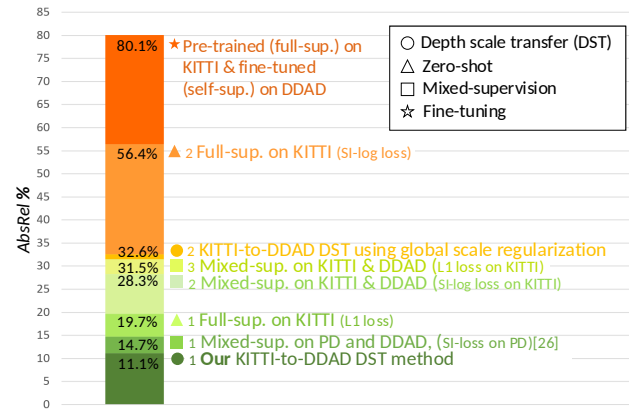}
	\end{center}
	\vspace{-35pt}
	\caption{Demonstrating various methods for achieving absolute depth predictions on the DDAD target dataset (front camera), when no GT depth labels from DDAD are available for training. The $AbsRel$ metric (lower is better) was measured on the DDAD validation dataset for each presented method. Mixed-supervised models ($\square$) were trained using self-supervision on both DDAD and another source dataset (KITTI or Parallel Domain (PD)) and with full-supervision on source GT depths using the mentioned loss.}
	\label{fig:figure1}
	\vspace{-15pt}
\end{figure}

Monocular depth estimation is a fundamental problem in computer vision with numerous scene understanding applications, such as autonomous driving and navigation \cite{ garg2016unsupervised,xiao2020multimodal,nidamanuri2021progressive,faria2021implementation}, robotics \cite{kim2021stereo,dong2022towards,shao2021self} and augmented reality \cite{diaz2017designing,kanbara2000stereoscopic}. Current methods for training Monocular Depth Estimators (MDE) include two major methodologies:  the first uses full-supervision \cite{fu2018deep,lee2019big,bhat2021adabins,guizilini2021sparse} or mixed-supervision \cite{kuznietsov2017semi,amiri2019semi,guizilini2020robust,baek2022semi}, where ground-truth (GT) depths are measured directly by LiDARs or reconstructed using a stereo setting \cite{kendall2017end,watson2019self,ranftl2020towards,cho2021deep}, achieving \emph{absolute} depth predictions. However, fine-tuning such models on new scenes or on images collected using different sensors, requires collecting their corresponding depth measurements, complicating the acquisition setup with additional depth sensors or cameras, increasing setup complexity and costs \cite{dong2022towards}. 

\begin{figure*}[htbp]
	\begin{center}
		\includegraphics[width=1.0\linewidth]{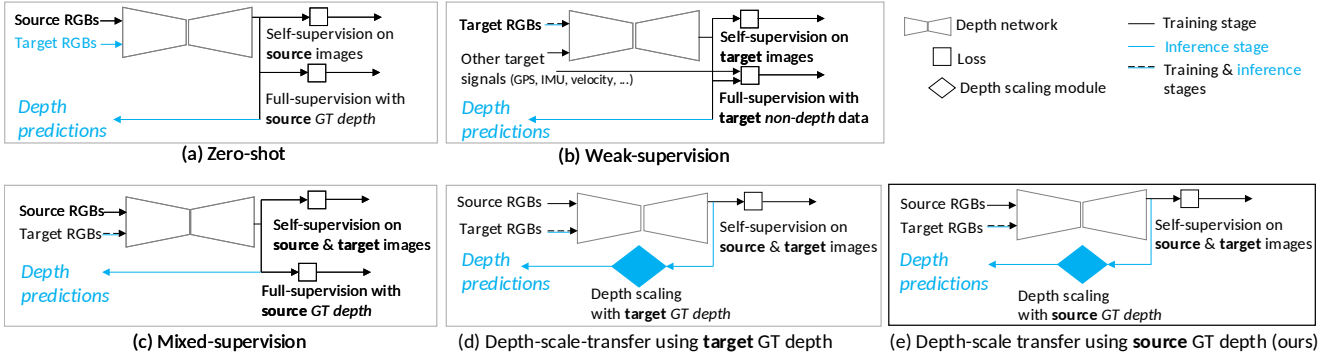}
	\end{center}
	\vspace{-15pt}
	\caption{Methods for inferring \emph{absolute} depth from single target RGB images, when no target GT \emph{depth} measurements are available.}
	\vspace{-10pt}
	\label{fig:depth_scale_modes}
\end{figure*}

Thus, to enable training or fine-tuning MDEs on a new domain using only images, multiple efforts  \cite{zhou2017unsupervised,godard2019digging,gordon2019depth,watson2021temporal,guizilini20203d,guizilini2022learning,chen2023frequency} were invested to improve the self-supervised regime. In this approach, two images acquired at different times are used to predict their depth using projective geometry \cite{hartley2003multiple}.
Due to the nature of this concept, the scale of the predicted depth is ambiguous \cite{hartley2003multiple}. In this work we refer to predicted depths that lack real-world scale units as \emph{up-to-scale} depths.

To compensate for the lack of GT depth measurements when collecting new data, one might suggest using already existing datasets collected with GT depths that do have real world-scale (\emph{source} dataset), and transferring from them the depth-scale property to the newly collected images (\emph{target} dataset). To show the motivation behind our work and the non-triviality of such depth-scale transfer between datasets collected using various sensors, we evaluated different approaches that might seem intuitive. To demonstrate a real-world scenario, we selected for the target domain the relatively new DDAD \cite{guizilini20203d} dataset and for the source domain we selected the well-known KITTI \cite{geiger2013vision} dataset, which was collected using a sensor with a different field-of-view (FOV). 

Our various early attempts included zero-shot estimation (see Figure \ref{fig:depth_scale_modes}a), fine-tuning using self-supervision on the target domain, and mixed-supervision on both domains (see Figure \ref{fig:depth_scale_modes}c). More details can be found in Sections \ref{previous_works} and \ref{results:naive_implementations}. However, Figure \ref{fig:figure1} shows that such methods result in poor accuracy.
A recent work \cite{guizilini2021geometric} applied mixed-supervision on the target DDAD images (without using any DDAD GT depths) and another source dataset, but did achieve high absolute depth accuracy on DDAD (see Figure \ref{fig:figure1}$\medsquare_1$). 

Thus, what impeded all our early attempts? 
An additional investigation indicated that the authors \cite{guizilini2021geometric} used for the source domain a synthetic dataset that was specifically generated with the same intrinsics as the target dataset \cite{PD}. However, this effort is costly and needs to be repeated for each sensor with a different FOV. Thus, in this work we study an \emph{alternative} for synthetic data generation for this purpose, by reusing existing synthetic or even \emph{real} datasets with GT depth labels, collected using \emph{variate} FOVs.

In order to use arbitrary datasets as source domains, we first studied fundamental aspects of the depth-scale, as predicted by self-supervised models trained using reprojected geometry.
We show that although such models can predict only up-to-scale depths, these are \emph{linearly} correlated with their respective GT depths, not only per a single image, but also across multiple images, displaying linear correlation characteristics per dataset, a property which we refer to in this work as \emph{linear} depth ranking.
Moreover, we show that when \emph{adjusting} images from two different domains to a single FOV, under the assumption of similar camera heights, training the MDE on images from both domains results in a shared depth ranking scale, regardless of possible domain gaps \cite{guo2018learning,zhao2019geometry}.

In this work we propose to transfer the depth-scale between domains by first training the MDE using self-supervision on images from both the source and target domain. Then, we model the relationship between the \emph{source} up-to-scale depths and their GT depth values using a \emph{single} scaling factor. Finally, we use this factor to scale the \emph{target} up-to-scale depth predictions, achieving real-world absolute depth predictions on the new domain.

\noindent In summary, our contributions are as follows:
\begin{itemize}
	\item {We propose a novel depth-scale transfer method that uses existing source data with GT labels, that does not increase the computations of the used MDE.}
	\item {The suggested method can reuse any existing datasets with GT labels, real or synthetic.}
	\item {Our method achieves comparable or better accuracy on KITTI, DDAD and nuScenes with respect to existing depth-scale transfer methods, even when using a vanilla lite MDE, as long as local motion is filtered out efficiently.}
	
\end{itemize}

\section{Related work}
\label{previous_works}

Over the years, various solutions were suggested to overcome the lack of target GT depth measurements for training MDEs to predict absolute depth from target images. Here we cover the main approaches, that are also presented by category in Figure \ref{fig:depth_scale_modes}. The first approach is implemented as zero-shot \cite{wang2018learning} (see Figure \ref{fig:depth_scale_modes}a), where a model is trained on source datasets, and used to infer depth on target images, in the hope of generalizing well on the new domain. A recent zero-shot model \cite{guizilini2023towards} successfully overcame the geometrical domain gap between the source and the target domain by training a transformer-based architecture on a variety of source datasets (containing more than 700,000  training images with GT) that were further augmented to support various focal lengths. In addition, the camera parameters were embedded to enable zero-shot capabilities on various target datasets. In our work, we show an alternative solution to close the geometrical domain gap that uses only few annotated \emph{source} samples (validation/test splits, less than 3,000 images) with a significantly lighter model (x50 less parameters). In addition, since our solution also uses target domain images, it could be re-adjusted to the new domain.

Another line of work used weak-supervision (see Figure \ref{fig:depth_scale_modes}b) to train the model directly on target data. Up-to-scale depth values were scaled during training with non-depth measurements such as the car velocity \cite{guizilini20203d}, its GPS location \cite{chawla2021multimodal} or IMU measurements \cite{zhang2022towards} through the relative translation regularization between frames in absolute distance units. However, such solutions also require signals acquisition with dedicated and costly setups. Other works \cite{guizilini2022full,wei2023surrounddepth} trained the MDE using self-supervision on target images from multiple cameras and regularized the estimated translation and rotation between the cameras using the measured pose translation and rotation between them, to indirectly regularize for the depth-scale. However, such solutions require the use of at least two cameras for collecting training data and rely on their calibration accuracy.

A variety of works used mixed-supervision (see Figure \ref{fig:depth_scale_modes}c) to transfer depth-scale from source datasets with GT depth labels to new target domains that were acquired without it. To overcome difficulties arising from training the MDE on source and target images collected by different sensors types, these solutions used for the source domain synthetic data that was specifically created with the same intrinsics, extrinsics and even similar geometrical structures as in the target domain. Thus, a large number of works were limited to demonstrating their method only on the target KITTI dataset, while the source data was taken from the synthetic vKITTI \cite{Gaidon:Virtual:CVPR2016} and vKITTI2 \cite{cabon2020vkitti2} that were specifically tailored to match KITTI. 
Since real and synthetic data differ in style, initial attempts focused on closing these gaps by incorporating style-transfer elements in the mix-supervision \cite{atapour2018real,zheng2018t2net,zhao2019geometry,lo2022learning}, achieving absolute depth predictions.
A recent work \cite{guizilini2021geometric} incorporated unsupervised domain adaptation principles in the mixed-supervision, successfully demonstrating depth-scale transfer to DDAD. To achieve this, the authors generated a synthetic source dataset \cite{PD} with the same intrinsics of the target domain. In our work we show an alternative for this step.

Recently, two approaches decoupled the up-to-scale depth ranking problem of the self-supervised regularization from the depth-scale estimation. In the first approach, the MDE was directly trained using self-supervision on target images. Then, depth-scaling was separately estimated per target \emph{image}, by using the known camera height and estimation of the road plane \cite{mccraith2020calibrating,xue2020toward} (see Figure \ref{fig:depth_scale_modes}d). However, this method requires sufficient visible free road during inference to estimate its plane, which is not always possible in traffic jams or turns. Our solution does not limit the target data by such condition.

In the second approach, the depth-scale is estimated by a module that only uses source data (see Figure \ref{fig:depth_scale_modes}e). A previous work \cite{bian2019unsupervised} suggested to train the MDE on the source data with self-supervision and an additional regularization loss to align the depth-scale of all predictions to a single arbitrary value. Then, the real depth-scale factor was separately estimated and corrected using GT depth measurements from a few source images. We also evaluated this approach when applying self-supervision on \emph{both} source and target domains using this regularization loss, and estimated the depth-scale from the source data, but did not achieve satisfactory results (see Figure \ref{fig:figure1}$\medcircle_2$ and Section \ref{results:naive_implementations}).

In a recent work \cite{swami2022you}, the authors trained an MDE using self-supervision on datasets from both source and target domains. Then, they trained a separate CNN module with full-supervision using source GT depths to predict the depth-scale of up-to-scale depth maps.
However, also this solution was demonstrated only on target datasets that have specially tailored synthetic source datasets. 

In our work we also propose to transfer the depth-scale from another source domain (see Figure \ref{fig:depth_scale_modes}e). However, contrary to previous solutions, we reuse any \emph{existing} source data type (real or synthetic), without limiting it to the intrinisics (FOV) of the target data. In addition, our depth-scale modality is implemented using a single scalar, thus does not increase the computations of the existing MDE.

\vspace{-4pt}
\section{Method}
\begin{figure*}
	\begin{center} 
		\includegraphics[width=0.95\linewidth]{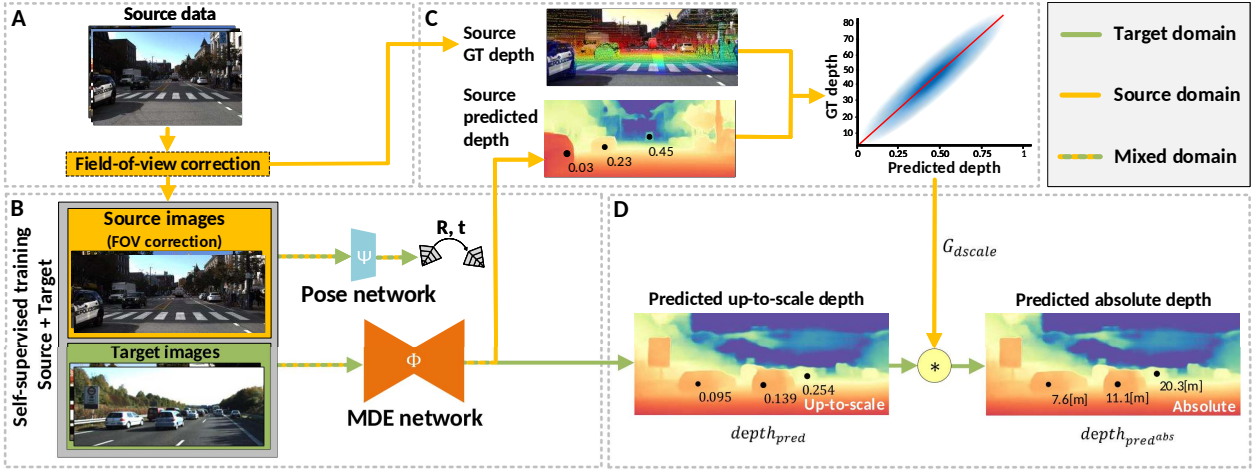}
	\end{center}
	\vspace{-15pt}
	\caption{Overview of our solution. (A) The FOV of source data is adjusted to the target FOV. (B) The depth and pose networks are trained using self-supervision on both source and target training images (mixed batches). 
		(C) Data from the source domain is used to generate the GT \vs predicted up-to-scale depths mapping. We apply the linear Theil-Sen regressor to calculate the $G_{dscale}$ depth-scaling factor for the trained MDE. (D) Estimated target up-to-scale depths are multiplied by the $G_{dscale}$ factor, resulting in absolute depth predictions.}
	\vspace{-10pt}
	\label{fig:architecture}
\end{figure*}

\subsection{Depth prediction architecture}
\label{methods:architecture}
To demonstrate the concept behind this work, we selected for the depth $\Phi$ and pose $\Psi$ networks similar designs as used for Monodepth2 \cite{godard2019digging}, consisting of 4.7M parameters. The MDE was trained using self-supervision with only a photometric loss \cite{godard2019digging}, after filtering local motion (see Supplementary material). Such architecture is sufficient for quantifying the effectiveness of the depth-scale transfer between domains. Additional depth accuracy improvements can be achieved using more complex architectures \cite{guizilini20203d,bhat2021adabins} or regularization losses \cite{guizilini2021geometric}, but are beyond the scope of this study.

\subsection{Adjusting the FOV of the source data to the target domain}
\label{methods:fov_correction}
Training an MDE on images collected using different camera FOVs (real or synthetic) introduces significant geometrical differences that a na\"ive self-supervised training regime cannot easily compensate for. 
To enable fine-tuning or training using self-supervision on images collected with sensors of different FOVs, without breaking the scene geometrical consistency, we propose to adjust the FOV of the source images to the FOV of the target images, resulting in training data with homogeneous FOV and aspect ratio. 
The FOV of a camera is calculated as:
\begin{equation}
	\vspace{-5pt}
	\angle{FOV}=2 \cdot atan\left(\frac{w}{2f}\right)
	\label{eq:FOV}
\end{equation}F))
where $f$ is the focal length of the camera and $w$ is the width of the image, both in pixel units. 
Let us denote the camera focal length and the image width in the target domain by $f_T$ and $w_T$ respectively, and the camera focal length in the source domain by $f_S$. To adjust the FOV of the source images to the target FOV, we center cropped the source image using the adjusted width $w_S$ to $w_{S \rightarrow T}$, as follows (assuming $f_S<f_T$): 
\begin{equation}
	\vspace{-5pt}
	\angle{FOV_T}=2 \cdot atan\left(\frac{w_T}{2f_T}\right)=2 \cdot atan\left(\frac{w_{S \rightarrow T}}{2f_S}\right)
	\label{eq:FOV_correction}
\end{equation}
resulting in:
\vspace{-10pt}
\begin{equation}
	\vspace{-0pt}
	w_{S \rightarrow T}=w_T\frac{f_S}{f_T}
	\label{eq:FOV_correction2}
\end{equation}	
\vspace{-1pt}
The adjusted crop height was similarly calculated $h_{S \rightarrow T}= h_T\frac{f_S}{f_T}$. Finally, the image crop was resized to the target image size to enable training on mixed batches. The same process was applied to the source GT maps. In case where $h_{S \rightarrow T}$ was higher than $h_T$ or $f_S$ was larger than $f_T$ the crop was padded, as detailed in the Supplementary material. 

\subsection{Estimating the depth scaling factor using a linear estimator}
\label{methods:global_estimator}

We inferred the trained MDE $\Phi$ on the \emph{source} test split and then analyzed the relationship between the predicted up-to-scale depths $depth_{pred}$ and their GT depth values. Since the training regime was designed to predict depths without any offset, we expected zero-depth predictions to match zero absolute depth.
The GT \vs the predicted up-to-scale depth scatter plot indicated a linear relationship (see Section \ref{results:linearity}) across the evaluated data, therefore we chose to model this relationship using a linear fitter with $G_{dscale}$ slope and zero intercept:
\vspace{-5pt}
\begin{equation}
	\vspace{-1pt}
	depth_{GT}\cong G_{dscale}\cdot depth_{pred}
	\label{eq:GT_to_pred_fitting}
\end{equation}

\noindent $G_{dscale}$ was estimated using the Theil-Sen regressor \cite{dang2008theil} (see Supplementary material), which was selected due to its high robustness to outliers. When the relationship was fitted per \emph{image}, the slope was noted with $I_{dscale}$.

A previous work \cite{swami2022you} chose to model the relationship between the source up-to-scale depth map and the depth-scale factor using a CNN-based network. For comparison purposes, we also implemented such solution. Additional details and analyses can be found in the Supplementary.

\subsection{Overview of our solution}
\label{methods:full_method}
\begin{table*}[t]
	
	\begin{center}
		\begin{tabular}{l|c|ccc|c|c|c}
			
			\toprule						
			&  			& \multicolumn{3}{c|}{\textbf{Our Source-to-Target DST}}      	 					& \textbf{Full-sup.} 			& \textbf{Self-sup.}   		& \textbf{Target-to-Target DST} 					\\ 
			{Target}						&  Source   & $\downarrow$${AbsRel}$    	&$\downarrow$$AbsRel_{norm}$    &$scale_{ratio}$    			   	 &$\downarrow$${AbsRel}$     			&  $\downarrow$$AbsRel_{norm}$       			 &$\downarrow$${AbsRel}$	\\ \hline\hline
			\multirow{ 2}{*}{DDAD1}      	&  KITTI  	& 0.111       		& 0.111 			&1.00±0.05							 & \multirow{ 2}{*}{0.100}  & \multirow{ 2}{*}{0.120}      		&\multirow{ 2}{*}{0.120}     	\\ 
			~      							&  vKITTI2  & 0.124       		& 0.127       		&1.01±0.05							 &         					&       							&	\\  \hline
			\multirow{ 2}{*}{KITTI}      	&  DDAD9  	& 0.084 			& 0.076 			&1.00±0.05  						 & \multirow{ 2}{*}{0.079}  & \multirow{ 2}{*}{0.087}      	    &\multirow{ 2}{*}{0.093}     	\\ 
			~      							&  vKITTI2  & 0.088       		& 0.075       		&1.02±0.06  						 &          				&        						    &	\\ \hline
			\multirow{ 2}{*}{nuScenes1}     &  KITTI  	& 0.165 				& 0.19 				&0.90±0.05  						 & \multirow{ 2}{*}{0.059}	& \multirow{ 2}{*}{0.332}      	& \multirow{ 2}{*}{0.425}       	\\ 
			~      							&  vKITTI2  & 0.141       		& 0.19       		&0.94±0.04  						 &          				&        						    &	\\

			\bottomrule
			
		\end{tabular}
	\end{center}
	\vspace{-15pt}
	\caption{Comparing the accuracy of our suggested depth-scale transfer (DST) method to other training regimes for the KITTI \cite{uhrig2017sparsity}, DDAD \cite{guizilini20203d} and nuScenes \cite{caesar2020nuscenes} target datasets. Our method is trained using self-supervision on a mix of images from the source and target datasets. In the last three columns show accuracy of models that are trained directly on the target train split using full-supervision, self-supervision and self-supervision and scaling using the $G_{dscale}$ calculated on the target train split.}
	\vspace{-10pt}
	\label{table:global_scale_transfer}
\end{table*}
First, we adjusted the FOV of the source domain data (train and test splits) to match the FOV of the target domain, as described in Section \ref{methods:fov_correction} and Figure \ref{fig:architecture}A. Training images from both source and target domains were randomly split into batches of four and used to train networks $\Phi$ and $\Psi$ in a self-supervised manner (see Section \ref{methods:architecture} and Figure\ref{fig:architecture}B).

Next, we used the depth network $\Phi$ to infer the up-to-scale depths of images from the test split of the \emph{source}. We fit the relationship between these predictions and their respective  GT depths (see Section \ref{methods:global_estimator} and Figure \ref{fig:architecture}C).

Finally, to estimate absolute depths on the $target$ domain, we inferred the up-to-scale depth of image $i$ from the target test split using the depth network $\Phi$ and multiplied the resulting up-to-scale depths by $G_{dscale}$, obtaining target absolute depth predictions $depth^{i}_{pred^{abs}}$ (see Figure \ref{fig:architecture}D):

\begin{equation}
	depth^{i}_{pred^{abs}}=G_{dscale}\cdot depth_{pred}^{i}
	\label{eq:dscale_factor}
\end{equation}

For comparison, we also trained the MDE using full-supervision with a L1 loss and \emph{target} GT depths to gauge the upper threshold accuracy of the used MDE architecture.

\subsection{Datasets}
\label{methods:datasets}

\textbf{KITTI}. This dataset \cite{Gaidon:Virtual:CVPR2016} is a common benchmark for depth evaluation. The front cameras have a FOV of 81\textdegree\ and are located 1.65 m above the ground.

\noindent \textbf{DDAD}. This dataset \cite{guizilini20203d} was collected using six cameras. The front and rear cameras have a FOV of 47\textdegree/83\textdegree\  and are located 1.55 m above the ground. Let us denote the data collected using the front and rear cameras as DDAD1 and DDAD9 respectively, following the dataset camera numbering convention.

\noindent \textbf{vKITTI2}. This dataset \cite{cabon2020vkitti2} was recently released as a more photo-realistic version of vKITTI \cite{Gaidon:Virtual:CVPR2016}. The camera has a FOV of 81\textdegree\  and is located 1.58 m above the ground. 

\noindent \textbf{nuScenes}. In our experiments we used only images from the front camera \cite{caesar2020nuscenes}, which has a FOV of 64.8\textdegree and located 1.51 m above the ground.

The KITTI, DDAD1 and nuScenes datasets were used as target domains. For the DDAD1 target domain we used KITTI and vKITTI2 as source domains. For the KITTI target domain we used vKITTI2, DDAD1 and DDAD9 as source domains (thus showing no limitation on the size of the source focal length) and for nuScenes1 we used KITTI and vKITTI2 as source domains. 
Additional details about these datasets can be found in the Supplementary material.

\subsection{Depth evaluation metrics}
\label{methods:metrics}
To measure the accuracy of the predicted absolute depth values, we used the absolute relative depth accuracy (\emph{AbsRel}) metric, without applying any normalization to the predicted depths \cite{eigen2014depth}.
To estimate the accuracy of the predicted up-to-scale depths, those were normalized first using the ratio between the medians of the predicted and the GT depth values (per image) \cite{zhou2017unsupervised}, resulting in $AbsRel_{norm}$ (see Supplementary material).

To assess scaling similarity to GT depth measurements, we also calculated the median ratio between the GT and the predicted absolute depths per image, and then averaged this value across the entire test split, resulting in $scale_{ratio}$.

\section{Results}
\label{results}	

Figure \ref{fig:figure1}$\medcircle_1$ and Table \ref{table:global_scale_transfer} show that our method was able to transfer the depth-scale from various domain sources (real or synthetic) acquired with different FOVs to various target datasets, achieving an $AbsRel$ of 11.1\% for the DDAD1 dataset, 8.4\% for the KITTI dataset and 14.1\% on the nuScenes1 dataset. The mean predicted depth deviated on average by less than 6\% with respect to the GT depth (see $scale_{ratio}$ in the fifth column of Table \ref{table:global_scale_transfer}), demonstrating the accuracy and effectiveness of our method.  

To estimate an upper bound accuracy of the used MDE architecture, we also directly trained the MDE with full-supervision using target GT depths. This training regime achieved for DDAD1 and KITTI an $AbsRel$ of 10.0\% and 7.9\% respectively (see Table \ref{table:global_scale_transfer}, sixth column). This shows that our depth-scale transfer method is competitive with fully-supervised methods that directly use target GT depth measurements, when controlling for the same MDE architecture. Additional depth-accuracy metrics, as well as visual examples of the estimated depth maps can be found in the Supplementary material.
\begin{table*}
	\centering
	
	\begin{tabular}{l|c|c|c|c|c|c|c} 
		\toprule
		
		{}		  &\multicolumn{3}{c|}{\textbf{All predictions}}      			& \multicolumn{4}{c}{\textbf{Predictions with $AbsRel_{norm}<15\%$}}  			\\ \cline{2-8}
		{Target}  &$I_{dscale}$ 	& $G_{dscale}$   		&$\uparrow$Pearson  &Remaining  		&$I_{dscale}$		& $G_{dscale}$    	& $\uparrow$Pearson   	\\ 
		{}		  &     			&						&coefficient   		&pixels  			&					&    	&			coefficient			\\ \hline\hline
		KITTI  	  & 85.7±9.3 		& 84.4					&0.93  				&83.6\%				&84.6±8.3			& 83.1				& 0.98					\\ 
		vKITTI2   & 105.2±16.1		& 107.5					&0.79  				&61.1\%				&115.2±11.1			& 113.4				& 0.98					\\
		DDAD1     & 124.2±14.1		& 122.5		    		&0.90  				&77.6\%				&125.9±12.3			& 124.4				& 0.98				    \\ 
		DDAD9     & 115.7±14.6		& 118.0		    		&0.85  				&59.9\%				&120.9±13.9			& 121.7				& 0.97				    \\ 
		nuScenes1 & 70.8±19.2		& 75.8		    		&0.76  				&53.0\%				&70.4±18.0			& 78.9				& 0.97				   	\\ 
		\bottomrule
		
	\end{tabular}
	\vspace{-5pt}
	\caption{Relationship between the GT and the predicted up-to-scale depths of our MDE, when separately trained on various datasets using self-supervision. Evaluations were done on their test split, using all predictions or predictions with $AbsRel_{norm}<15\%$ (remaining pixels \% after the filtering is mentioned in the fifth column). For each data slicing the relationship was separately fitted \emph{per image} using the linear model described in Section \ref{methods:global_estimator}. Its mean and standard deviation across all test images is reported in the $I_{dscale}$ column.
		The relationship was also fitted using the same model directly on \emph{all} test images and depicted by the $G_{dscale}$ factor and the Pearson correlation coefficient.}
	\vspace{-5pt}
	\label{table:single_domain_results}	
	\label{table:SFM_linearity} 
\end{table*}

The $AbsRel_{norm}$ metric is agnostic to the depth-scale estimation quality, thus reflects errors related to other factors such as the limited capacity of the network, insufficient loss regularization, poor image quality and domain gaps due to limited training data, \etc. The results in Table \ref{table:global_scale_transfer} (third \vs fourth columns) indicate of a small difference between the $AbsRel$ and the $AbsRel_{norm}$, providing an additional metric to quantify our depth-scale related error.

To evaluate the impact of training the MDE on data from an additional (source) domain on the depth estimation accuracy of the target domain, we also trained the MDE using self-supervision directly on the target domain. The results in Table \ref{table:global_scale_transfer}  (fourth \vs seventh column) indicate that training on additional data did not deteriorate the accuracy of the model. This was specifically prominent on the nuScenes1 dataset, where training on an additional source of data significantly improved the $AbsRel_{norm}$ metric from 33.3\% to 19\%, which was also previously reported as significantly poor \cite{guizilini2023towards}. We refer to the reasons behind this significant improvement in the Supplementary material. 

For comparison, we also scaled the self-supervised predictions using the estimated global $G_{dscale}$ calculated on the target data, achieving similar results to the scaling per image using GT target measurements (Table \ref{table:global_scale_transfer}, seventh \vs eight columns).
In the next sub-sections we will present the analyses that laid the foundations for our suggested inter-domain depth-scale transfer method.

\subsection{Studying per-domain the relationship between GT and predicted up-to-scale depths}
\label{results:linearity}

\begin{figure}[h]
	\includegraphics[width=1.0\linewidth]{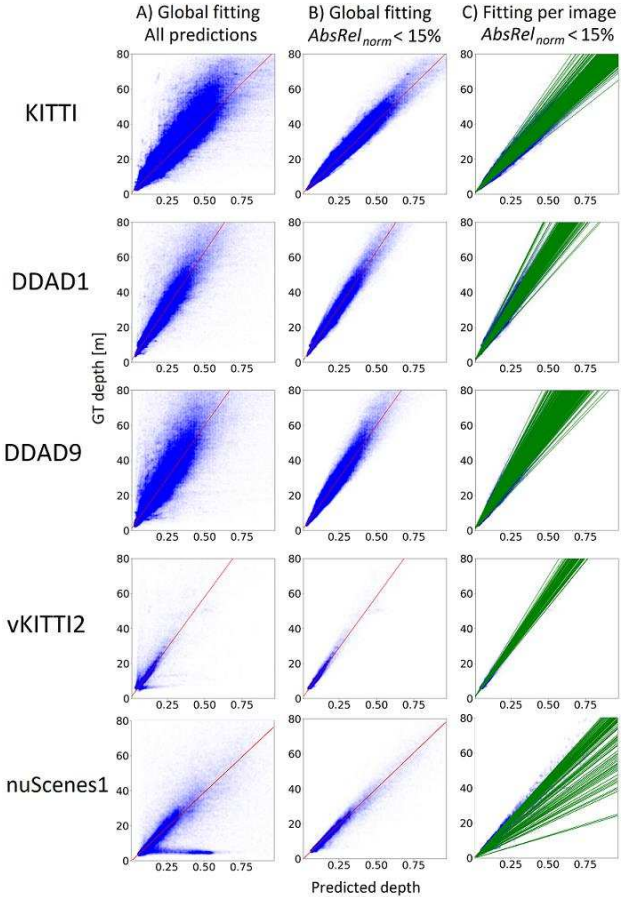}
	\vspace{-20pt}
	\caption{Scatter plots of the GT \vs the predicted up-to-scale depth values. Our MDE was separately trained on various datasets using self-supervision and inferred on test images from the same domain. (A) All test data. The red line depicts the linear fit applied on all data (see Eq. (\ref{eq:GT_to_pred_fitting}). (B) Predictions with $AbsRel_{norm}>15\%$ were filtered out and the fitting was recalculated. (C) The GT \vs the predicted depth relationship was similarly linearly fitted \emph{per image} and depicted by different green lines.}
	\label{fig:SFM_linearity}
	\vspace{-5pt}
\end{figure}

We started our analysis by separately training the MDE on datasets from various domains. For each trained model the relationship between the GT and the predicted up-to-scale depths was analyzed on its respective test split.
An MDE trained using projective geometry \cite{hartley2003multiple,eigen2014depth} is expected to linearly rank the predicted depth per \emph{image} \cite{zhou2017unsupervised}. During training, no additional bias corrections were applied to the network output, thus we expected GT values close to zero to be mapped to predicted up-to-scale depths close to zero (\ie zero offset) (see Figure \ref{fig:SFM_linearity}A). 
For each test split we fit \emph{per image} the GT to up-to-scale depths using the linear model described in Section \ref{methods:global_estimator} (${I}_{dscale}$ in Table \ref{table:SFM_linearity} and Figure \ref{fig:SFM_linearity}C).

A more thorough analysis of the GT \vs the predicted up-to-scale scatter plot in Figure \ref{fig:SFM_linearity}A indicated of a common linear trend across all test images, achieving a Pearson correlation coefficient bigger than 0.76 (see Table \ref{table:single_domain_results}). Measurements were fitted using the linear model described in Section \ref{methods:global_estimator} and the calculated $G_{dscale}$ factor is reported in Table \ref{table:SFM_linearity}.
The scatter plots in Figure \ref{fig:SFM_linearity}A revealed that some measurements are outside of the main linear trend. We hypothesized that these outliers could result from poor up-to-scale predictions of the MDE. To validate this assumption, we filtered out predicted depths with $AbsRel_{norm}>$15\%. As seen in Figure \ref{fig:SFM_linearity}B, the filtering indeed removed the majority of outliers, increasing Pearson correlation coefficient to above 0.97. 
Although the obtained slope might slightly vary across images (see Figure \ref{fig:SFM_linearity}C), their mean is comparable to the slope $G_{dscale}$ calculated across the entire split (see Table \ref{table:single_domain_results}), suggesting that the training process of the MDE converges the depth ranking  per-image into a \emph{common} linear ranking scale for all depths in the scenes of the domain it was trained on.

We hypothesize that the observed ${I}_{dscale}$ variability could be explained by imperfect convergence due to non-optimal training losses, architecture capacity, image  quality, generalization gaps and other factors.
Conducting experiments to measure the impact of each factor on the slope variability across images is beyond the scope of this work; however, Table \ref{table:single_domain_results} shows that when removing poor up-to-scale predictions ($AbsRel_{norm}>15\%$), the variability decreased, supporting this assumption.

To reinforce that the observed linearity results from the projective geometry based loss, regardless of the network architecture, we also analyzed up-to-scale predictions of a significantly different architecture, PackNet \cite{guizilini20203d}. This network  uses 3D convolutions and a different encoder-decoder architecture, but was trained with a similar loss. As shown in the Supplementary material, the predicted up-to-scale depths of this self-supervised MDE are also linearly correlated to the GT depths.

The obtained mean of the slopes per individual image is similar to $G_{dscale}$ and their variability is substantially smaller than their mean. Thus, in this work we evaluated how well the single depth-scaling factor $G_{dscale}$  per dataset could be used to model the relationship between the GT and the predicted up-to scale depths.

\subsection{Training the MDE on images from two domains}
\label{results:mixed_training}

Modeling the GT to predicted up-to-scale depth relationship resulted in highly varying $G_{dscale}$ values for different domains (see Table \ref{table:single_vs_mixed_training}, third column). Thus, even if a self-supervised MDE is able to linearly rank depths, the modeled $G_{dscale}$ is not transferable to other domains. 
To demonstrate that $G_{dscale}$ differences cannot be explained only by different FOVs, an additional MDE was separately trained on source datasets after adjusting the source images to match the target FOV (see Table \ref{table:single_vs_mixed_training}, fourth column). Applying the source $G_{dscale}$ on the target, achieved an $AbsRel$ of 0.225/0.337/0.491/0.123 for each row in Table \ref{table:single_vs_mixed_training}, showing that this step on its own is insufficient.

To overcome the $G_{dscale}$ high differences between each pair of domains, we hypothesized that training the MDE using self-supervision jointly on both the source and the target datasets could potentially result in ranking the depths of both domains on a \emph{common} scale, thus achieving a \emph{de facto} inter-domain depth ranking. 
To enable training on two domains, the images from the source domain were adjusted to the FOV of the target domain, as described in Section \ref{methods:fov_correction}. 
The fifth column in Table \ref{table:single_vs_mixed_training} shows that training the MDE on data from two domains resulted in similar $G_{dscale}$ values for both the source and the target test splits, suggesting that the calculated $G_{dscale}$ on the source data could be applied also on the target data. 
Finally, we also evaluated the effect of incorrect source image resizing to align the size of source and target image without considering the effect of FOV, we implemented two na\"ive alternatives. More details about this analysis can be found in the Supplementary material.

\begin{table}
	
	\centering
	\begin{tabular}{l|c|c|c|c} %
		
		\toprule					
		\multicolumn{2}{c|}{} 					 					 &\multicolumn{3}{c}{\textbf{$G_{dscale}$} ($AbsRel_{norm}<$15\%)}  				\\ \cline{3-5}
		\multicolumn{2}{c|}{} 					 					 &{Separate}  			&{Separate}							&{Joint}				\\
		\multicolumn{2}{c|}{} 					 					 &{trainings}  			&{trainings}						&{training}				\\
		\multicolumn{1}{c|}{\textbf{Test domain}}&\textbf{Type} 	 &{}					& {S$\xrightarrow{\text{FOV}}$T}	&{S$\xrightarrow{\text{FOV}}$T}				\\ \hline\hline
		KITTI									 &T   				 &83.1					&83.1								&99.3					\\ 
		DDAD9									 &S  				 &121.7					&102.8								&99.5					\\  \hline	
		
		KITTI									 &T   				 &83.1					&83.1								&100.7					\\ 
		vKITTI2									 &S  				 &113.4					&113.4								&106.5					\\  \hline
		
		DDAD1									 &T   				 &124.4					&124.4								&136.5					\\ 
		KITTI									 &S  				 &83.1					&63.9								&136.1					\\  \hline
		
		DDAD1									 &T   				 &124.4					&124.4								&127.3					\\ 
		vKITTI2									 &S  				 &113.4					&127.6								&129.2					\\
		\bottomrule
		
	\end{tabular}
	\vspace{-5pt}
	\caption{$G_{dscale}$ values estimated on test splits of various datasets (first column). Second column indicates which dataset was used as source (S) or target (T). Third column: two MDEs were trained separately on the source and the target training datasets; Fourth column: two MDEs were trained separately on the source and the target training datasets, the source images were adjusted to the target FOV (S$\xrightarrow{\text{FOV}}$T); Fifth column: a single MDE was trained on a mixture of training data from the target and the source domains after the source images were adjusted to the target FOV.}
	\label{table:single_vs_mixed_training}
	\vspace{-10pt}
\end{table}

\subsection{Straightforward attempts for depth-scale-transfer between domains}
\label{results:naive_implementations}	

We report the accuracy of various early approaches that we implemented in attempt to compensate for the lack of GT target measurements, as mentioned in Section \ref{intro}.
We started with a zero-shot approach, where the MDE was trained using full-supervision with varying losses (L1, SI-log) on the KITTI dataset. The model was inferred on the DDAD1 dataset, but achieved non-satisfactory results (see Figure \ref{fig:figure1}$\triangle$), as expected when the geometry of the target scene is significantly different. We also evaluated the accuracy of a fully-supervised model on KITTI, that was fine-tuned using self-supervision on the DDAD1 target. However, Figure \ref{fig:figure1}$\medstar$ shows poor accuracy, due to the loss of depth-scale. 

For the mixed-supervision implementation (see Figure \ref{fig:depth_scale_modes}c), we did not have access to the specially tailored synthetic data (PD \cite{PD}) that the authors \cite{guizilini2021geometric} used as source data for the target DDAD1 dataset, thus we used as a source the KITTI dataset. However, this approach obtained unsatisfactory results (see Figure \ref{fig:figure1}$\medsquare_{2,3}$), since retrospectively, these datasets cannot be straightforwardly mixed together due to their different geometry (as supported by results in Table \ref{table:single_vs_mixed_training}).

We also trained the MDE using self-supervision on both the source and target train splits, with an additional regularization loss that aligns the depth-scale of predictions in both domains \cite{bian2019unsupervised}, and then estimated the depth-scale factor using GT depths of a few source images (category (e) in Figure \ref{fig:depth_scale_modes}). However,  this approach also failed to achieve satisfactory accuracy (see Figure \ref{fig:figure1}$\medcircle_2$). This low accuracy can be also explained by the inability to straightforwardly mix together datasets with different FOVs.

\subsection{Accuracy comparison to other methods}

Finally, we compared our method to other absolute depth predictors that were trained without any target GT depth measurements (see Table \ref{table:others}), using the $AbsRel$ metric on KITTI, DDAD1 and nuScenes1 target domains. These were described in Section \ref{previous_works} and Figure \ref{fig:depth_scale_modes}a-e and include zero-shot \cite{guizilini2023towards}, weakly-supervised methods \cite{guizilini20203d,chawla2021multimodal,zhang2022towards,guizilini2022full}, mixed-supervised methods \cite{lo2022learning,guizilini2021geometric} and depth-scale transfer from target GT depth \cite{mccraith2020calibrating,xue2020toward} or source GT depth \cite{swami2022you}.

The zero-shot model \cite{guizilini2023towards} achieved better accuracy on two out of the three target domains, however it used a significantly bigger training dataset (700,000 images) and larger architecture (232.6M parameters \vs 4.7M in our model). Nonetheless, our suggested method achieved competitive or better accuracy than the rest of the methods, while using both existing real or synthetic source datasets, collected with different FOVs than the target data, without being limited by road visibility.

\setlength\tabcolsep{2pt} 
\begin{table}
	\centering
	
	
	\begin{tabular}{l|c|c|c|c|c|c}
		\toprule					
		\textbf{Target} 			&\textbf{Method}				&\textbf{Ref.}					&\textbf{Source}  	&\multicolumn{3}{c}{$\downarrow$${AbsRel}$}										\\ 
		&								&								&  					&\cite{eigen2014depth} 		&\cite{guizilini20203d}	&\cite{caesar2020nuscenes}	\\ 		\hline\hline	
		~\multirow{ 14}{*}{KITTI}	&{Zero-shot}	 				&\cite{guizilini2023towards}	&Multiple			&0.100  					&-						&-   					\\ \cline{2-7}	
		~							&\multirow{3}{*}{Weak-sup}		&\cite{guizilini20203d}			&-					&0.107						&-					  	&-						\\  
		~							&								&\cite{zhang2022towards}		&-				    &0.108						&-						&-    					\\
		~							& 								&\cite{chawla2021multimodal}	&-					&0.109						&-					  	&-						\\  \cline{2-7}
		~							&\multirow{4}{*}{Mixed-sup}	 	&\cite{lo2022learning}			&CS					&0.119   					&-						&-   					\\ 
		~							& 								&\cite{lo2022learning}			&vKITTI				&0.120						&-						&-   					\\ 	
		~							& 								&\cite{guizilini2021geometric}	&vKITTI2			&0.107						&-						&-   					\\  \cline{2-7}
		~							&{DST-target} 					&\cite{mccraith2020calibrating}	&-  				&0.113					 	&-						&-   					\\ 
		~							& 								&\cite{xue2020toward}			&-  				&0.118						&-						&-   					\\ \cline{2-7}
		~							& \multirow{4}{*} {DST-source}	&\cite{swami2022you}			&vKITTI2 			&0.109						&-						&-   					\\
		~							& 								&\textbf{Ours}					&vKITTI2			&0.108						&-						&-   					\\ 	
		~							&								&\textbf{Ours}					&{DDAD9}			&0.110						&-						&-   					\\ 	
		~							&								&\textbf{Ours}					&{DDAD1}			&0.117						&-						&-   					\\ \hline	
		\multirow{ 4}{*}{DDAD1}		&{Zero-shot}					&\cite{guizilini2023towards}	&Multiple			&-							&0.100					&-						\\	\cline{2-7} 	
		~							&{Weak-sup} 					&\cite{guizilini2022full}		&-  				&-					 		&0.130					&-   					\\ 	\cline{2-7}
		~							&{Mixed-sup}	 				&\cite{guizilini2021geometric}	&PD 				&-							&0.147  				&-						\\  \cline{2-7} 						
		~							&\multirow{2}{*}{DST-source}	&\textbf{Ours}					&KITTI  			&-							&0.111					&-   					\\  
		~							&~ 								&\textbf{Ours}					&vKITTI2			&-							&0.124    				&-   					\\  \hline 	
		~\multirow{ 5}{*}{nuSc1}	&\multirow{1}{*}{Zero-shot}	 	&\cite{guizilini2023towards}	&Multiple			&-   						&-						&0.150   				\\  \cline{2-7} 	  	
		~							&{Weak-sup} 					&\cite{guizilini2022full}		&-  				&-					 		&-						&0.186   				\\ 	\cline{2-7}
		~							& \multirow{2}{*}{DST-source}	&\textbf{Ours}					&KITTI				&-							&-						&0.165					\\ 	
		~							&								&\textbf{Ours}					&vKITTI2			&-							&-						&0.141					\\ 	
		
		\bottomrule
		
	\end{tabular}
	\vspace{-5pt}
	\caption{Comparing absolute depth estimation accuracy of methods that do not have access to target GT depth measurements during training. nuSc1, CS and PD are abbreviations of nuScenes1, Cityscapes and Parallel Domain. References in the $AbsRel$ columns indicate the used evaluation dataset. Training regimes abbreviations: weak-supervision (Weak-sup), mixed-supervision (Mixed-sup), depth-scale transfer using source-data (DST-source), depth-scale transfer using target-data (DST-target).}
	\label{table:others}
	\vspace{-10pt}
\end{table}

\section{Discussion}
 
 Training or fine-tuning absolute depth estimators on collected images from a different domain without using their GT depth measurements is an existing challenge with real-world significance. Current solutions include zero-shot models that are trained on a large number of labeled images, mixed-supervised solutions that require generating tailored synthetic data to resemble the scenes of the new domain or the target sensors specification, or solutions that estimate the depth-scale from the target domain using some apriori knowledge about the camera setup.
 
 In this work we suggested an alternative method that reuses a relatively small number of existing labeled real or synthetic images, to transfer from them the depth-scale to new domains, emphasizing the insensitivity of this approach to style or structural domain gaps.
 Our method leveraged the observed depth ranking linearity of projective geometry self-supervision, resulting in a lite-weight solution that does not alter the MDE architecture. 
 
 Although we demonstrated the method using a self-supervised vanilla architecture and efficient local motion filtering, we achieved competitive or better results than other existing solutions. 
 More advanced architectures and losses can independently improve the up-to-scale depth predictions, which in turn impact the accuracy of the absolute depth predictions, even when using an ideal depth-scaling factor. We postulate that more accurate self-supervised MDEs can also reduce the variability of the GT to up-to-scale relationship across images, thus enabling a better generalization of the single depth-scale correction factor.
 
 Finally, the suggested solution is also highly applicative for continuous learning, where depth models require adjustments to new scenes, without the need to collect additional GT depth measurements.
  
{
    \small
    \bibliographystyle{ieeenat_fullname}
    \bibliography{egbib}
}


\end{document}


\maketitle

\section{Methods}

\subsection{Training methodology additional details}
\label{methods:additional_details}

Training an MDE using self-supervision aims at reconstructing the view from an image collected at time $t\pm1$ from an image collected at time $t$, using the estimated depth from the image at time $t$ and the estimated relative pose between the frames \cite{zhou2017unsupervised,godard2019digging}. Let us denote by $\Phi$ a network that receives as input an image $I_t$ and by $d_t$ its estimated up-to-scale depth output, and by $T_{t\rightarrow t\pm1}$ the relative pose between two frames $I_t$ and $I_{t\pm1}$ estimated by a network $\Psi$.
Given the camera intrinsics $K$, a homogeneous pixel $p_t\in I_t$ can be mapped to a homogeneous pixel $p_{t\pm1} \in I_{t\pm1}$ as follows \cite{hartley2003multiple}:
\begin{equation}
	\hat{p}_{t\pm1}~\sim {K} \hat{T}_{t\rightarrow t\pm1}d_t(p_t) {K}^{-1}p_t
	\label{eq:1} 
\end{equation}

In this work we adopted the depth $\Phi$ and pose $\Psi$ network architectures presented in Monodepth2 \cite{godard2019digging}, but replaced the backbone of the depth network $\Phi$ with MobileNetV2 \cite{sandler2018mobilenetv2} instead of the original ResNet18 \cite{he2016deep}.
The MDE was trained using a phototometric loss \cite{godard2019digging} for 15 epochs, with a learning rate of $10^{-4}$ and for another five epochs with a learning rate of $10^{-5}$.

In the original work \cite{godard2019digging}, the authors used the depth network $\Phi$ to estimate the inverse depth, then they inverted the prediction and scaled it to depth values of [0.1, 80] per image to avoid shrinking of the estimated depth. This correction introduces a bias, which we avoided in this work, by directly predicting up-to-scale depths. The predicted up-to-scale depth range is between 0 and 1, as determined by the last sigmoid activation of the $\Phi$ decoder. We demonstrated that this implementation achieves similar accuracy, while providing simpler interpretability, which was used for additional analyses, as shown in the Results section in the main text.

\subsection{Local motion filtering}
\label{methods:lmf}

Monocular depth estimators (MDEs) trained using self-supervision from image pairs collected at different times are highly sensitive to local motion \cite{godard2019digging}, which breaks the scene stationarity assumption on such regions, invalidating Eq. \ref{eq:1}.
\begin{figure}
	\begin{center}
		\includegraphics[width=1.0\linewidth]{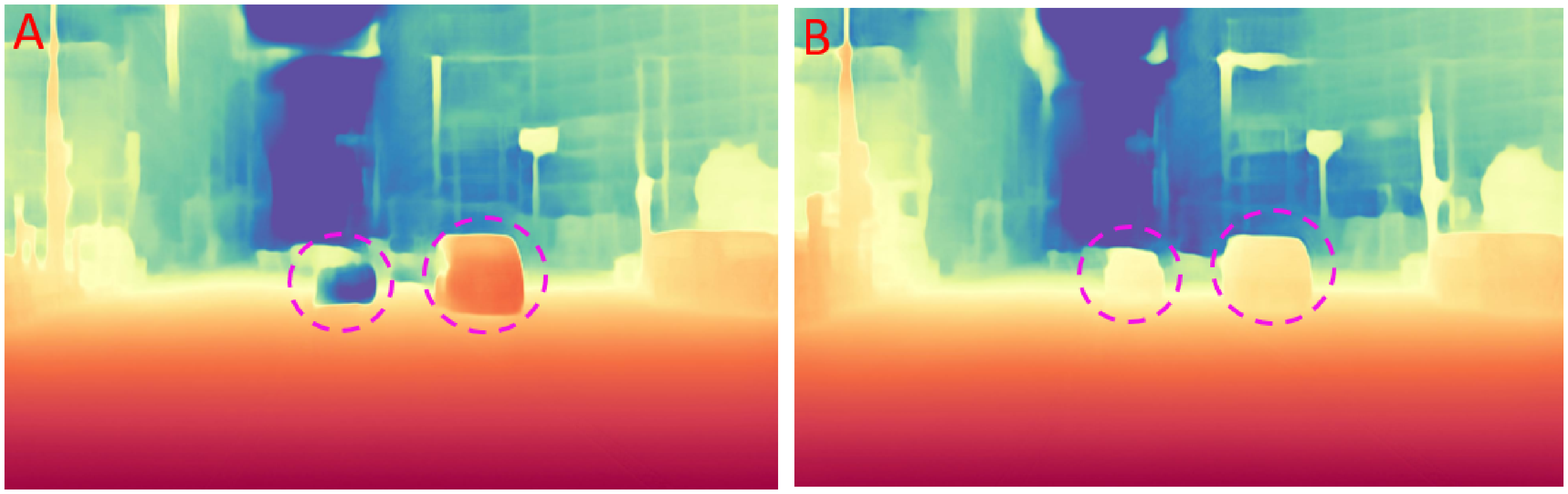}
	\end{center}
	\vspace{-15pt}
	\caption{Demonstrating the effect of local motion on self-supervised training (inferred image taken from the DDAD1 dataset). A. The resulting depth estimation map when the MDE was trained using self-supervision with limited local motion filtering \cite{godard2019digging}. B. The depth estimation map after applying our additional local motion masks on potentially moving vehicles during training. Dashed circles are located on moving cars in opposite directions (left/right car is moving from/towards the ego-car).}
	\label{fig:local_motion}
	\vspace{-10pt}
\end{figure}

Previous efforts to correct this phenomenon included filtering vehicles that moved with the same velocity as the ego-motion \cite{godard2019digging}. Others regularized the loss by using additional optical flow predictions \cite{guizilini2022learning}, applied separate regularization for static and dynamic objects \cite{casser2019depth,li2021unsupervised} or fully-supervised these regions with GT depth \cite{watson2021temporal}.

In this work, we suggested to filter non-stationary vehicles from the training data by directly exploiting this self-supervised training weakness. 
Specifically, we calculated the relative depth on dynamic objects, such as cars, with respect to stationary segments such as the road. On moving objects the relative depth should significantly differ when predicted by a self-supervised MDE that is sensitive to local motion \vs an MDE that is fully-supervised and does not suffer from this weakness.

To this end, we first trained the MDE using self-supervision as described in Section \ref{methods:additional_details}. We used this network to infer up-to-scale depths on all the training dataset, creating $d_{unsup}^i$ estimations ($i$ represents the index of an image). Then, we trained the same depth estimation network $\Phi$ using full-supervision with a simple L1 loss (see Section \ref{methods:full_supervision}) on the \emph{source} training dataset that has available GT depth measurements. We used this network to infer the depth of images from the source and the target training datasets, creating $d_{sup}^i$ estimations. The fully-supervised training is not strictly required and any fully-supervised model can be used, as long as it can generalize well enough to produce continuous depth maps on roads and cars (without necessarily resulting in correct absolute depth values).

We then utilized a semantic-segmentation network \cite{tao2020hierarchical} to detect the road region and scaled both $d_{sup}^{i}$ and $d_{unsup}^{i}$ by dividing them with their respective \textit{road-pixels} median, resulting in $d_{sup-scaled}^i$ and $d_{unsup-scaled}^i$ estimations.

Next, we employed an off-the-shelf instance-level segmentation network to estimate the location of vehicles \cite{wu2019detectron2}. For each vehicle we calculated the absolute difference $|d_{sup-scaled}^i - d_{unsup-scaled}^i|$ values, setting a local-motion cutoff value above $C$ for at least $R\%$ of each vehicle pixels. 
In this work we empirically set $C$ to 1.5 (reflecting $d_{unsup-scaled}^{i}$ depth deviations of more than 50\% with respect to  $d_{sup-scaled}^i$) and $R$ to 10 (reflecting that at least 10\% of pixels are affected by local motion).

We repeated the self-supervised training of models $\Phi$ and $\Psi$ using the additionally created local-motion masks.
These step reduced areas that violated the static scene assumption, which in turn improved poor estimations on moving objects. An example of the resulting local motion filtering applied during training is shown in Figure \ref{fig:local_motion}.

It is worth mentioning that local motion masks are applied only on the training data to enable correct photometric loss only on static objects. During inference the depth is inferred on single full images, that are not filtered by local motion masks. Therefore, no assumption regarding road visibility is required during the inference stage (as used in the training phase).

\subsection{Estimating the depth-scale factor}
In this work we fitted the relationship between the predicted up-to-scale depths of the MDE and their GT absolute depth values using the Theil-Sen regressor \cite{dang2008theil}, which is robust to outlayers (see main Text, Figure 4A). The slope of this estimator is calculated as the median of the slopes of all lines that are defined by each two points.

\subsection{Estimating the absolute depth using full-supervision}
\label{methods:full_supervision}
To enable comparison of our absolute depth predictions to predictions of a similar architecture trained in a fully-supervised manner, the depth network $\Phi$ was trained directly on images and GT depth measurements from the \emph{target} train dataset, using an L1 loss between the predicted and the GT depth values \cite{kendall2017end}.
The model was trained for 15 epochs with a learning rate of $10^{-4}$, and for another five epochs using a learning rate of $10^{-5}$. 

\subsection{Estimating the per-image depth scaling factor using an alternative CNN network}
\label{methods:dscale_net}
A recent work \cite{swami2022you} also suggested a depth-scale transfer architecture that uses source GT depths. In their work the depth scale correction module was implemented using convolutional network that receives as input features from the last layer of the depth decoder network and was trained to estimate the depth-scale factor \emph{per image}.
To compare our global estimation approach to a depth scaling factor \emph{per image} approach, we also implemented a convolutional neural network $\Gamma$. The global scaling model receives as input the up-to-scale depth map, thus for a fair comparison, network $\Gamma$ was also designed to receive as input up-to-scale depth maps. We implemented the network using a MobileNetV2 architecture \cite{sandler2018mobilenetv2} and converted its head into a regression head. The network was trained to regress for \emph{each} input up-to-scale depth map $depth_{pred}^{i}$ its depth-scale correction factor $net_{dscale}^{i}$, such that:

\begin{equation}
	depth^{i}_{pred^{abs}}=net_{dscale}^{i}\cdot depth_{pred}^{i}
	\label{eq:image_dscale_factor}
\end{equation}
Up-to-scale depth maps were inferred on images from the source train dataset and used as inputs to train network $\Gamma$. The network was trained for 15 epochs in a fully-supervised manner using L1 loss between the predicted $depth^{i}_{pred^{abs}}$ (see Eq. (\ref{eq:image_dscale_factor})) and the absolute depth $depth_{GT}^i$ values, with a learning rate of  $10^{-4}$. We used the \emph{source} test dataset to select the epoch with the lowest $AbsRel$ (see Section \ref{methods:metrics}) and reported the selected model accuracy on the \emph{target} test dataset.

\subsection{Depth evaluation metrics}
\label{methods:metrics}	
We calculated the absolute relative depth metric (\emph{AbsRel}) as previously described \cite{eigen2014depth}:
\begin{equation}
	Abs Rel=\frac{1}{T}\sum^{T}_{t}\frac{1}{N_t}\sum^{N_t}_{n}(|\frac{depth_{pred^{abs}}^{t,n} - depth_{GT}^{t,n}}{depth_{GT}^{t,n}}|)
	\label{eq:ARE_abs}
\end{equation}
where $n$ denotes pixels with valid GT depths, $N_{t}$ the number of pixels with valid GT depth in image $t$ and $T$ the number of images in the test dataset. 

To measure the \emph{AbsRel} of predicted up-to-scale depths, predicted up-to-scale depths of image $t$ were first normalized using the ratio $\alpha$ between the median of the predicted and the GT depth values (per image) \cite{zhou2017unsupervised}, resulting in $AbsRel_{norm}$.

\begin{equation}
	Abs Rel_{norm}=\frac{1}{T}\sum_{T}^{t}\frac{1}{N_t}\sum^{N_t}_{n}(|\frac{\alpha depth_{pred}^{t,n} - depth_{GT}^{t}}{depth_{GT}^{t}}|)
	\label{eq:ARE_rel}
\end{equation}

where $\alpha=\frac{median(depth_{GT}^{t})}{median(depth_{pred}^t)}$.

To assess scaling similarity we also calculated the median ratio between the predicted absolute depths and the GT depths per image, and then averaged this value across the entire test dataset, resulting in $scale_{ratio}$:

\begin{equation}
	scale_{ratio}=\frac{1}{T}\sum ^{T}_{t}median(\frac{depth_{pred^{abs}}^{t}}{depth_{GT}^{t}})
	\label{Scale_similarity}
\end{equation}

\subsection{Datasets}
\label{methods:datasets}

\textbf{KITTI}. This dataset \cite{geiger2013vision} is considered the standard benchmark for depth evaluation. Its Eigen split \cite{eigen2014depth} resulted in 39,810 training images, 4,424 validation and 697 evaluation images.
This dataset was recorded during various days and the stereo rectification resulted in slight FOV variability  between 81.43\textdegree\  and 82.68\textdegree. 
The cameras were located 1.65 m above the ground. 

To enable comparison to previous works, we evaluated our method on the 697 Eigen split evaluation images (see main text, Table 5). In the rest of the paper we  reported accuracy on the newer Eigen Benchmark evaluation dataset \cite{uhrig2017sparsity}, which has an improved ground truth depth and contains 652 evaluation images. To ease the reading flow, we refer to the evaluation dataset as the test dataset.
The FOVs of datasets that were used as a source domain to KITTI were adjusted to 81.43\textdegree.

\noindent \textbf{DDAD}. This depth evaluation benchmark \cite{guizilini20203d} was collected from various locations in the world using six cameras (front, rear, sides). 
The training dataset contains 12,560 images and the validation dataset contains 3,950 images (per camera). To ease the reading flow, in this work we refer to the validation dataset as our test dataset. 
The front camera has a FOV between 47.66\textdegree\ and 48.26\textdegree, while the rear camera has a FOV between 82.21\textdegree\ and 84.46\textdegree. All cameras were located 1.49-1.54 m above the ground. We refer to data collected using the front and the rear cameras as DDAD1 and DDAD9 respectively, following the dataset camera numbering convention. The FOVs of all datasets that were used as a source domain to DDAD1 were adjusted to 47.85\textdegree.

\noindent \textbf{nuScenes}. For training and evaluation we used images from the front camera \cite{caesar2020nuscenes}. Evaluation was done on the official evaluation split of the front camera, which has 6019 a FOV of 64.8\textdegree and located 1.51 m above the ground. For evaluation we applied the Garg crop as used in KITTI \cite{eigen2014depth}.

\noindent \textbf{vKITTI2}. This synthetic dataset \cite{cabon2020vkitti2} was recently released as a more photo-realistic version of vKITTI \cite{Gaidon:Virtual:CVPR2016}, containing reconstructions of five sequences found in the KITTI odometry benchmark with full-pixel ground truth. We used only camera 0 from all scenes and all of its rotations $\pm15$\textdegree and $\pm30$\textdegree, from the clear day simulation, resulting in 9,560 images. From each sequence and rotation angle we used the first 90\% of images for training and the rest for testing purposes.
Similarly to KITTI, the FOV of the images in this dataset is 81.16\textdegree, created using a camera located $\sim$1.58 m  above the ground.

In this work we defined KITTI, DDAD1 and nuScenes1 as target datasets. For the DDAD1 target domain we used KITTI and vKITTI2 as source domains. For the KITTI target domain we used vKITTI2, DDAD1 and DDAD9 as source domains. For the nuScenes target domain we used KITTI and vKITTI2 as source domains.
When the applied FOV adjustment (see Section 3.2 in the main text) resulted in crops larger than the original image (\eg KITTI FOV correction to DDAD1 or nuScenes) or when the source focal length is smaller than the target focal length (\eg DDAD1 FOV correction to KITTI), these sections were filled using reflection zero/padding for RGB images/GT depth. In experiments where KITTI was defined as the target domain, input images for the various trained networks were resized to 320x1024. When DDAD1 was defined as the target domain, the input images for the various trained networks were resized to 608x960. When nuScenes1 was defined as the target domain, training images were resized to 488x800. We cap the depth range to 80 m during evaluation for both KITTI, DDAD1 and nuScenes1 target datasets.

\section{Results}

\subsection{The effect of poor adjustment of source images}

To enable training on mixed batches, our FOV adjustment method resizes source images to match the size of target images. To evaluate the effect of incorrect source image resizing, we implemented two na\"ive alternatives that do not consider the FOV of images: in the first option (A), source images were centered to target images and cropped to match their size; in another alternative (B), source images were resized to the target images size (width-wise), and then were cropped to match the target height. We applied these methods when using DDAD as a source domain to the KITTI target domain, and then repeated the self-supervised training using mixed batches, calculated the $G_{dscale}$ and applied depth-scale transfer. The results in Table \ref{table:FOV_ablation} show that although the $AbsRel_{norm}$ is comparable to $AbsRel_{norm}$ of our FOV adjustment method, the $AbsRel$ is significantly higher. This indicates that resizing source images without taking into consideration the target FOV, fails to enable correct depth-scale transfer using the suggested method.

\begin{table}
	
	\begin{adjustbox}{width=\columnwidth,center}
		\begin{tabular}{l|c|c|c|cc} %
			\toprule					
			
			\textbf{Resizing} 		&\textbf{Type} 	&\textbf{Domain}	&\textbf{Slope}	 &$\downarrow$${AbsRel}$    			& $\downarrow$$AbsRel_{norm}$		\\
			\textbf{method} 		& 				&					&	 			 &    						& 						\\ \hline\hline	
			\multirow{ 2}{*}{A}		&S				&DDAD9   			&97.08			 &\multirow{2}{*}{0.423} 	&\multirow{2}{*}{0.094}	\\ 
			&T				&KITTI  			&70.51			 &							&						\\  \hline	
			
			\multirow{ 2}{*}{B}		&S				&DDAD1   			&163.63			 &\multirow{2}{*}{0.847} 	&\multirow{2}{*}{0.084}	\\ 
			&T				&KITTI  			&86.69			 &							&						\\  
			
			\bottomrule
			
		\end{tabular}
		
	\end{adjustbox}
	\vspace{-10pt}
	\caption{Alternative source image resizing implementations to match the target image size. Second column indicates the used source (S) and target (T) datasets for training on both domains. The slopes are calculated using the $G_{dscale}$ model.}
	\label{table:FOV_ablation}
	\vspace{-10pt}
\end{table}

\begin{figure}
	\includegraphics[width=1.0\linewidth]{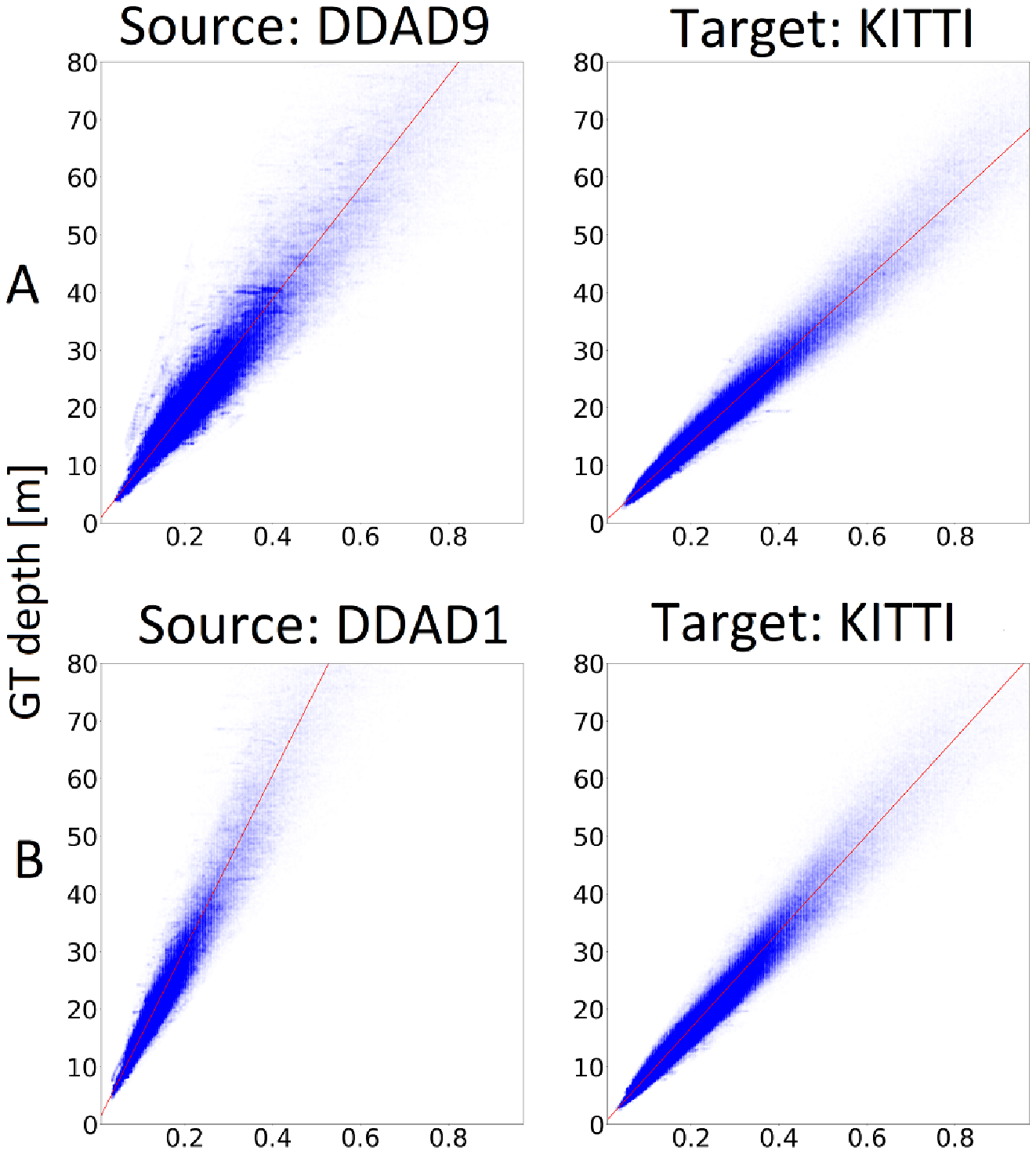}
	
	\caption{Na\"ive resizing of source images to the size of target images using the two different approaches (A) and (B) described in Section 4.4. Up-to-scale depths were inferred using an MDE that was trained using self-supervision on batches with mixed source and target images. Source images were adjusted using the suggested na\"ive resizing methods. GT to up-to-scale depth scatter plots were calculated on the source and the target test datasets. }
	\label{fig:crop_ablation}
\end{figure}

Figure \ref{fig:crop_ablation} shows the obtained GT to up-to-scale depth scatter plots after self-supervised training on mixed batches of images from the target and source domain (after applying the na\"ive resizing methods on the source images).

\subsection{Depth estimations on nuScenes}
Table 1 in the main text shows that when training the MDE on nuScenes data mixed with another dataset (KITTI or vKITTI2), the $AbsRel_{norm}$ decreased from  33.2\% to 19\%, as also depicted in the estimated depth maps depicted in Figure \ref{fig:nuscenes}. An additional investigation indicated that the nuScenes train dataset includes images that were captured while raining, thus blurring parts images used for training or contained night scenes that included strong light reflections. Such poor estimations are also depicted in Figure 4 (main text) in the form of a long tail that deviates from the general linear trend (red line).

We postulate that training in combination with cleaner datasets such as KITTI or vKITTI2 enabled the model to improve depth predictions on such regions, thus decreasing overall prediction error.

\begin{figure}
	\begin{center}
		\includegraphics[width=1.0\linewidth]{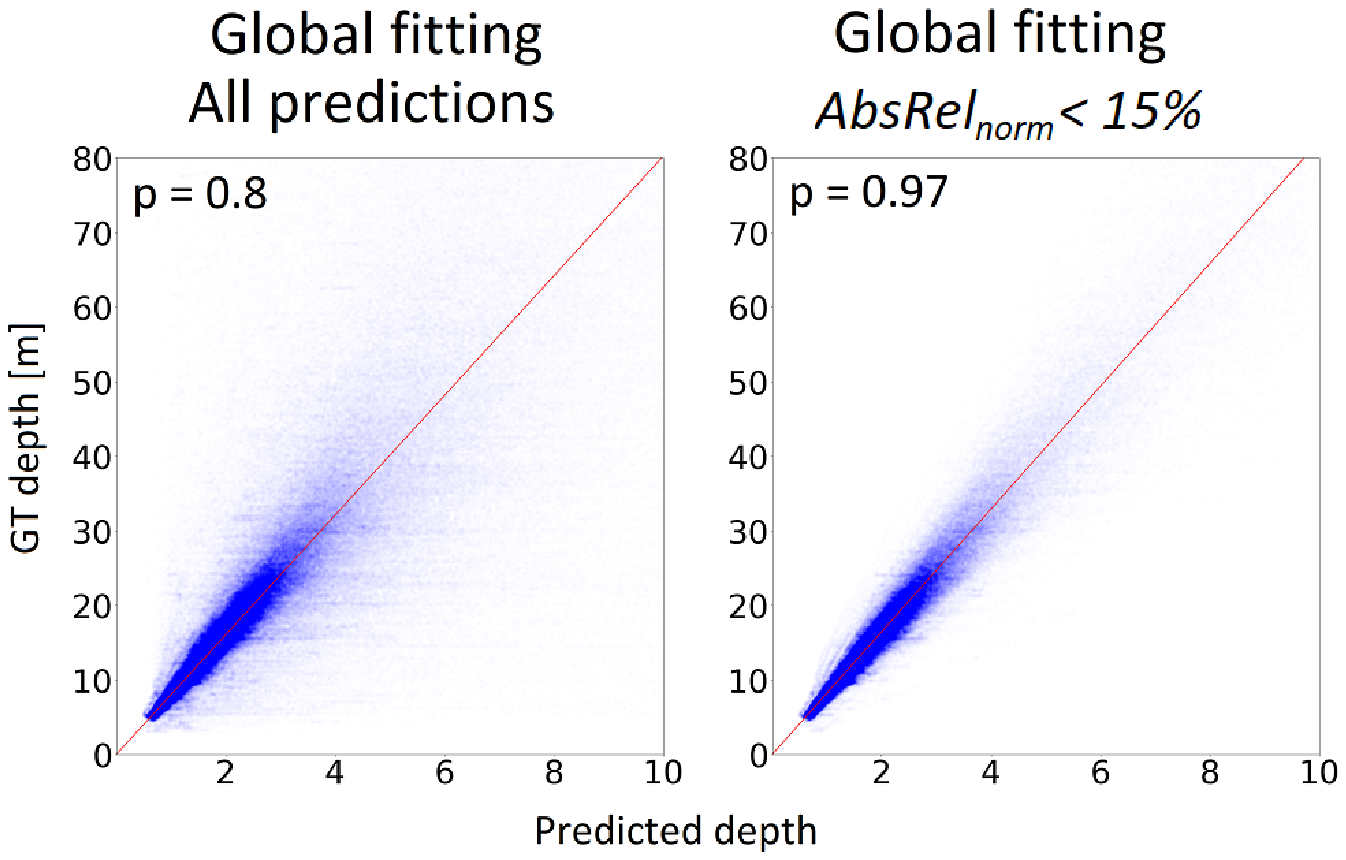}
	\end{center}
	\vspace{-15pt}
	\caption{GT to up-to-scale predicted depth estimations using the PackNet architecture. Left: all predictions. Right: predictions with $AbsRel_{norm}<15\%$. The fitting line was calculated using Eq. (5) in the main text and depicted with a red line. The calculated Pearson coefficient (p) was added for each scatter plot.}
	\label{fig:packnet}
	\vspace{-10pt}
\end{figure}

\subsection{Demonstrating depth ranking linearity of another depth estimation architecture}
\label{results:packnet_linearity}
We hypothesized that the depth linear ranking property observed in self-supervised MDEs is independent of the used depth estimation architecture, but determined by the used training loss. To this end, we analyzed depth predictions of the PackNet architecture \cite{guizilini20203d}. This architecture utilizes 3D convolutions and significantly differs from our encoder-decoder architecture, but uses a similar training loss as our self-supervised training regime. In their work, the authors predicted 1/depth, thus the up-to-scale depth values are not limited to 1.

We inferred depth predictions on the DDAD1 dataset using the publicly available PackNet model, which was trained using self-supervision on 384x640 resized images from the same domain.
Figure \ref{fig:packnet} demonstrates the linear relationship between the GT and predicted up-to-scale depths of this network, achieving a Pearson coefficient of p=0.8 on all predictions and p=0.97 on predictions with $AbsRel_{norm}<15\%$, reinforcing our assumption.

\subsection{Transferring depth scale from the source to the target domain - additional metrics}
\label{results:scale_transfer_results}

We report in Table \ref{table:global_scale_transfer_metrics} additional accuracy metrics of our depth-scale transfer method.
We also evaluated the per-image depth-scale network $\Gamma$ described in Section \ref{methods:dscale_net} and reported its accuracy metrics in Tables  \ref{table:dscalenet_transfer_metrics}.
The results show that when given the same input information, the $\Gamma$ network that has 3.5M parameters achieved similar or lower accuracy as our depth-scale transfer method implemented using a single scalar $G_{dscale}$ per MDE. 

Figures \ref{fig:F_S_KITTI_DDAD9}-\ref{fig:F_S_DDAD1_vKITTI2} show inferred up-to-scale depths from the source and the target domain, when separately training the MDE using self-supervision on each one of the domains (upper row) and up-to-scale depths after training the MDE on mixed-batches from both the source and the target domains (bottom row). FOV of source images in both cases was adjusted to the target FOV. Self-supervised training on mixed batches from different domains was shown to not qualitatively degrade the estimated depth maps.

\begin{figure*}
	\begin{center}
		\includegraphics[width=0.95\linewidth]{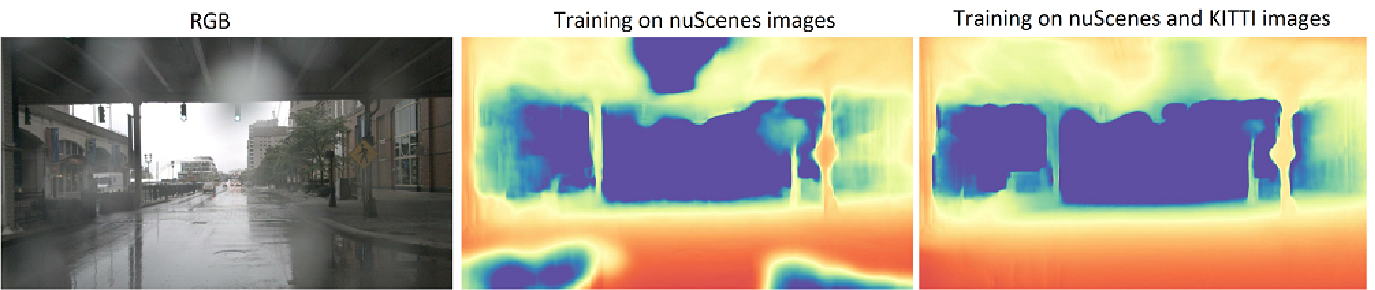}
	\end{center}
	\caption{Depth estimations on nuScenes images. Left: an RGB image from nuScenes. Middle: estimated depth of the image in left using the MDE trained with self-supervision on nuScenes training split. Right: estimated depth using the MDE trained on both nuScenes and KITTI training split.}
	\label{fig:nuscenes} 
\end{figure*}

{\small
	\bibliographystyle{ieeenat_fullname}
	\bibliography{egbib_suppl}
}

\begin{table*}
	\centering
	
	\begin{adjustbox}{width=1\textwidth}
		\begin{tabular}{ll|ccccc|ccc|c}
			\toprule						
			{}								& {}    	&\multicolumn{5}{c|}{Lower is better {$\downarrow$}}      	&\multicolumn{3}{c|}{Higher is better {$\uparrow$}}  	&1 is better \\ \cline{3-11}
			\\[-1em]
			{Target}						&  {Source} & ${AbsRel}$      	& $AbsRel_{norm}$    & Sq Rel 	& RMSE 		& RMSE$_{log}$		&$\delta<1.25$  & $\delta<1.25^{2}$  & $\delta<1.25^{3}$   	&$Scale_{ratio}$\\ \hline \hline
			\multirow{2}{*}{DDAD1}      	&  KITTI  	& 0.111      		& 0.111 			 & 1.310    & 6.769	    &0.185				& 0.875			& 0.964				 & 0.984				&1.00±0.05		\\ 	
			&  vKITTI2  & 0.124       		& 0.127     		 & 1.506    & 7.061	    &0.196				& 0.858			& 0.958				 & 0.982				&1.01±0.05		\\ \hline 
			\multirow{2}{*}{KITTI}      	&  DDAD9  	& 0.084 			& 0.076 			 & 0.403 	& 3.459	   	&0.124				& 0.925			& 0.987		 		 & 0.997				&1.00±0.05 		\\ 
			&  vKITTI2  & 0.088      		& 0.075     		 & 0.409 	& 3.393	    &0.124				& 0.927			& 0.989			 	 & 0.997				&1.02±0.06		\\

			\bottomrule 
			
		\end{tabular}
	\end{adjustbox}
	\caption{Our depth-scale-transfer method using the $G_{dscale}$ linear fitter. For KITTI, metrics were calculated on the Eigen benchmark evaluation data \cite{uhrig2017sparsity}.}
	\label{table:global_scale_transfer_metrics}
	
	\vspace{10mm}
	
	\begin{adjustbox}{width=1\textwidth}
		\begin{tabular}{ll|ccccc|ccc|c}
			\toprule						
			{}							& {}    	&\multicolumn{5}{c|}{Lower is better {$\downarrow$}}      	&\multicolumn{3}{c|}{Higher is better {$\uparrow$}}  	& 1 is better\\ \cline{3-11}
			\\[-1em]
			{Target}					&  {Source} & ${AbsRel}$      	& $AbsRel_{norm}$    	& Sq Rel 		& RMSE 		& RMSE$_{log}$		&$\delta<1.25$  & $\delta<1.25^{2}$  & $\delta<1.25^{3}$ &$Scale_{ratio}$\\ \hline \hline
			\multirow{2}{*}{DDAD1}   	&  KITTI  	& 0.135      		& 0.118 			    & 1.554 	    & 7.405	    & 0.208				&0.844			& 0.955				 &0.982				 &0.94±0.06		\\ 	
			&  vKITTI2  & 0.169      		& 0.134       		    & 2.200 		& 8.693	    & 0.224				&0.802			& 0.947			     &0.979				 &1.08±0.07		\\  \hline 
			\multirow{2}{*}{KITTI}      &  DDAD9  	& 0.091 			& 0.075 				& 0.44 		    & 3.665		& 0.132				&0.913			& 0.986				 &0.997				 &0.97±0.06		\\
			&  vKITTI2  & 0.101      		& 0.076       			& 0.513 	    & 3.601	    & 0.132				&0.922			& 0.988				 &0.997				 &1.05±0.06		\\

			\bottomrule
		\end{tabular}
	\end{adjustbox}
	\caption{Depth-scale-transfer implemented using the per-image depth-scale, implemented using network $\Gamma$. For KITTI, metrics were calculated on the Eigen benchmark evaluation data \cite{uhrig2017sparsity}.}
	\label{table:dscalenet_transfer_metrics}
\end{table*}

\clearpage
\begin{figure*}[t]
	\begin{center}
		\includegraphics[width=1.0\linewidth]{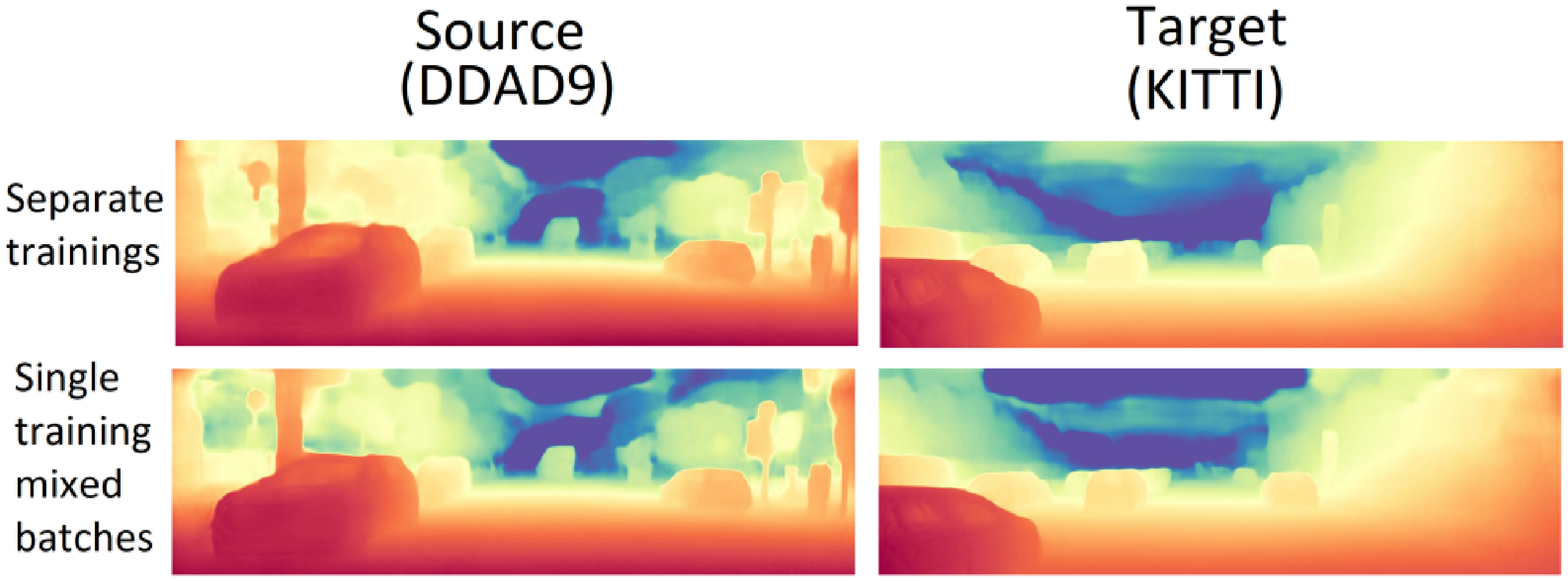}
	\end{center}
	\caption{Upper row: two MDEs are separately trained using self-supervision on DDAD9 and KITTI.
		Bottom row: One MDE is trained using self-supervision on mixed batches from DDAD9 and KITTI. In both experiments the FOV of DDAD9 images was adjusted to the KITTI FOV.}
	\label{fig:F_S_KITTI_DDAD9}
\end{figure*}

\begin{figure*}
	\begin{center}
		\includegraphics[width=1.0\linewidth]{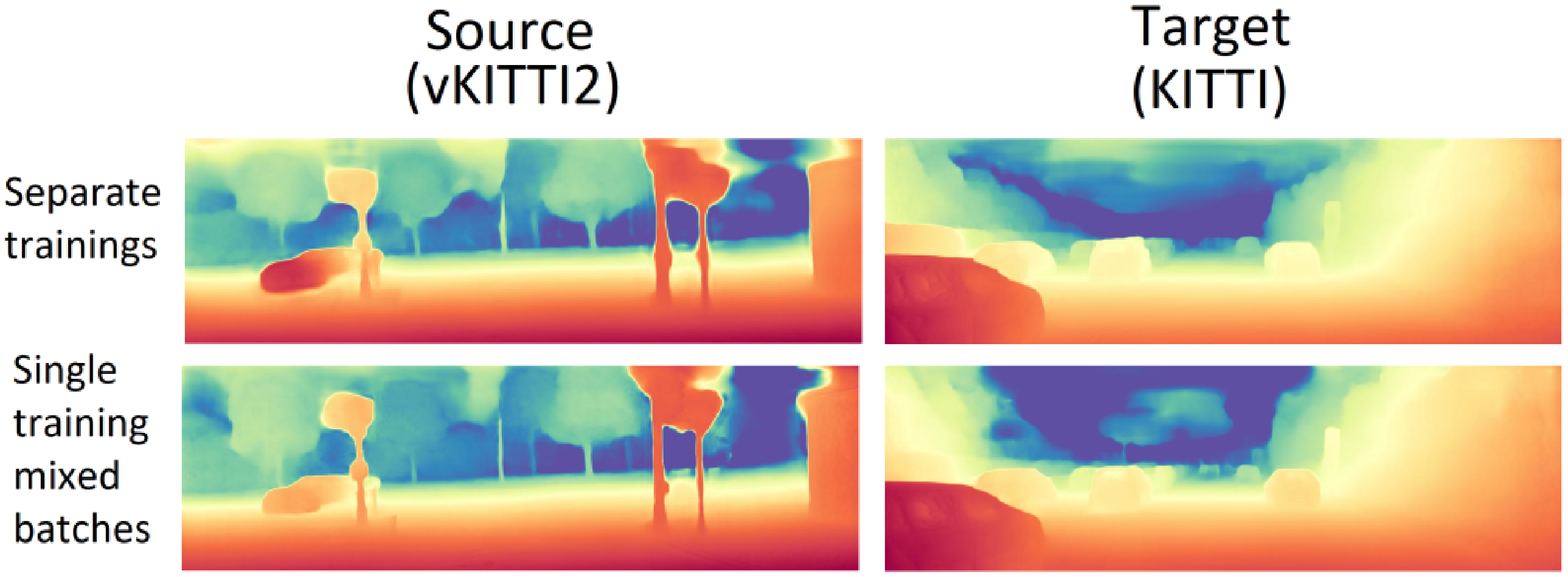}
	\end{center}
	\caption{Upper row: two MDEs are separately trained using self-supervision on vKITTI2 and KITTI.
		Bottom row: One MDE is trained using self-supervision on mixed batches from vKITTI2 and KITTI.}
	\label{fig:F_S_KITTI_vKITTI2} 
\end{figure*}

\begin{figure*}
	\begin{center}
		\includegraphics[width=0.75\linewidth]{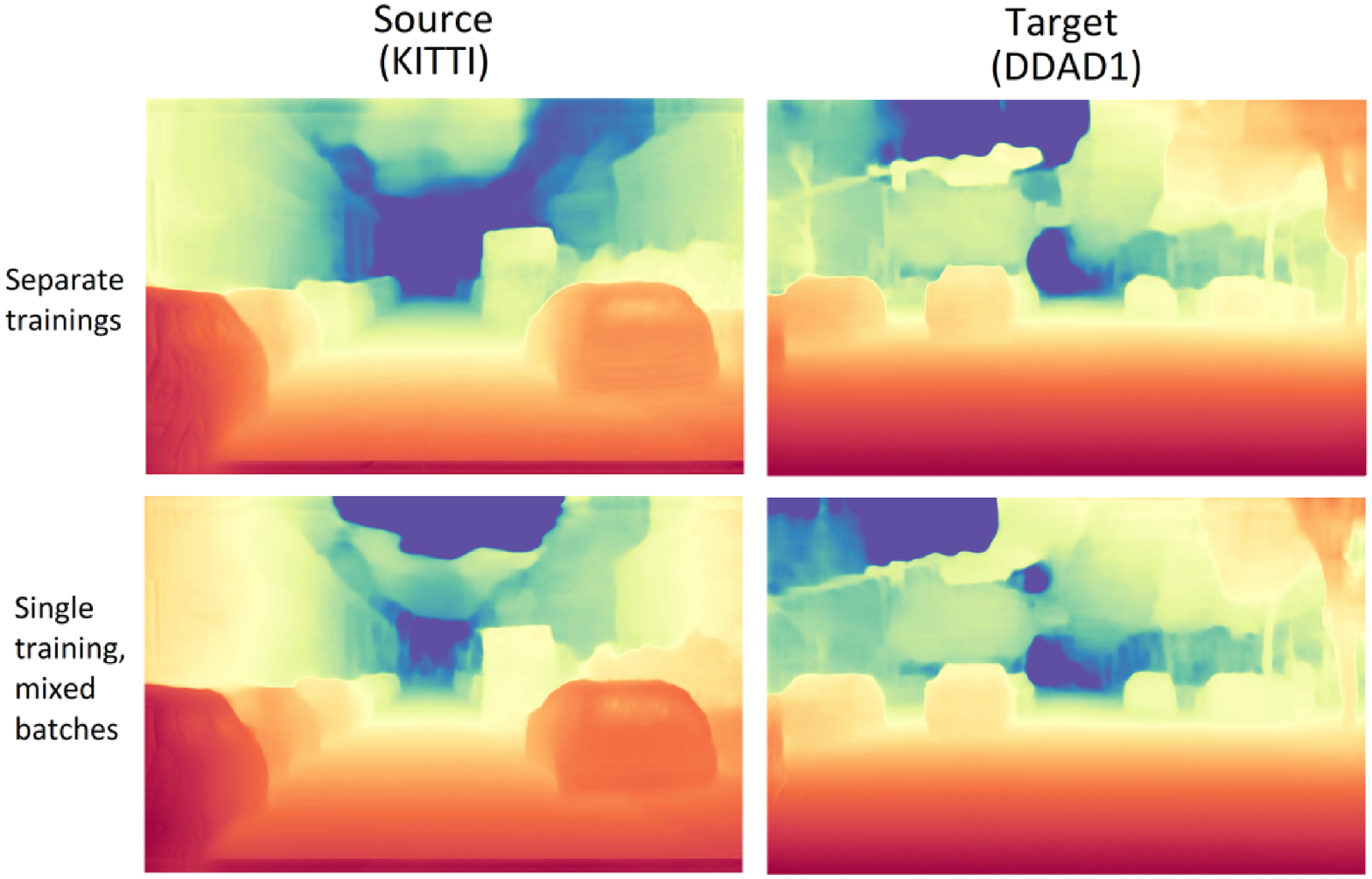}
	\end{center}
	\caption{Upper row: two MDEs are separately trained using self-supervision on KITTI and DDAD1.
		Bottom row: One MDE is trained using self-supervision on mixed batches from KITTI and DDAD1. In both experiments the FOV of KITTI images was adjusted to the DDAD1 FOV.}
	\label{fig:F_S_DDAD1_KITTI} 
\end{figure*}

\begin{figure*}
	\begin{center}
		\includegraphics[width=0.75\linewidth]{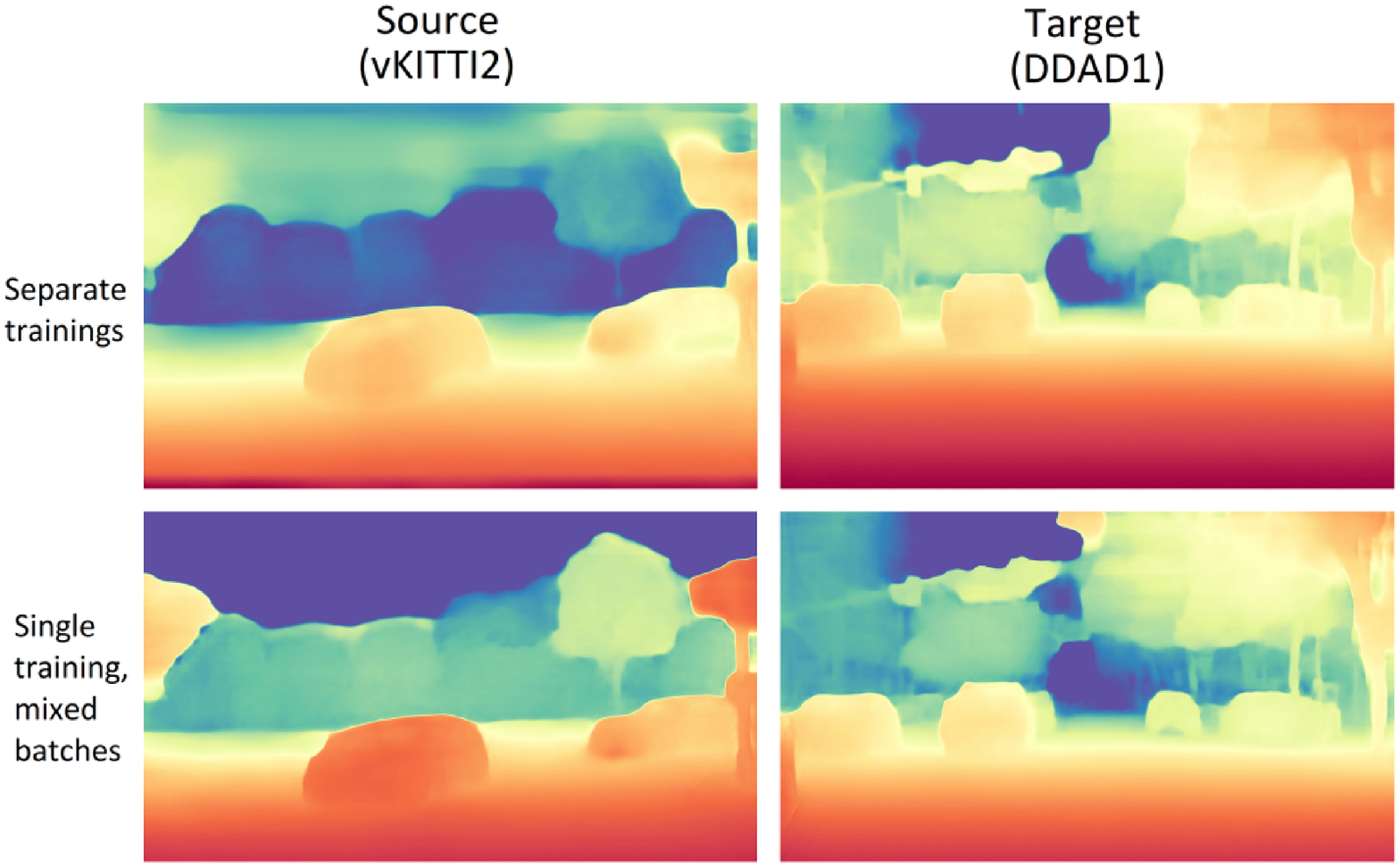}
	\end{center}
	\caption{Upper row: two MDEs are separately trained using self-supervision on vKITTI2 and DDAD1.
		Bottom row: One MDE is trained using self-supervision on mixed batches from vKITTI2 and DDAD1. In both experiments the FOV of vKITTI2 images was adjusted to the DDAD1 FOV.}
	\label{fig:F_S_DDAD1_vKITTI2} 
\end{figure*}


%% file: preamble.tex
\usepackage[dvipsnames]{xcolor}


%% file: main.bbl
\begin{thebibliography}{53}
\providecommand{\natexlab}[1]{#1}
\providecommand{\url}[1]{\texttt{#1}}
\expandafter\ifx\csname urlstyle\endcsname\relax
  \providecommand{\doi}[1]{doi: #1}\else
  \providecommand{\doi}{doi: \begingroup \urlstyle{rm}\Url}\fi

\bibitem[Amiri et~al.(2019)Amiri, Loo, and Zhang]{amiri2019semi}
Ali~Jahani Amiri, Shing~Yan Loo, and Hong Zhang.
\newblock Semi-supervised monocular depth estimation with left-right
  consistency using deep neural network.
\newblock In \emph{2019 IEEE International Conference on Robotics and
  Biomimetics (ROBIO)}, pages 602--607. IEEE, 2019.

\bibitem[Atapour-Abarghouei and Breckon(2018)]{atapour2018real}
Amir Atapour-Abarghouei and Toby~P Breckon.
\newblock Real-time monocular depth estimation using synthetic data with domain
  adaptation via image style transfer.
\newblock In \emph{Proceedings of the IEEE conference on computer vision and
  pattern recognition}, pages 2800--2810, 2018.

\bibitem[Baek et~al.(2022)Baek, Kim, and Kim]{baek2022semi}
Jongbeom Baek, Gyeongnyeon Kim, and Seungryong Kim.
\newblock Semi-supervised learning with mutual distillation for monocular depth
  estimation.
\newblock In \emph{2022 International Conference on Robotics and Automation
  (ICRA)}, pages 4562--4569. IEEE, 2022.

\bibitem[Bhat et~al.(2021)Bhat, Alhashim, and Wonka]{bhat2021adabins}
Shariq~Farooq Bhat, Ibraheem Alhashim, and Peter Wonka.
\newblock Adabins: Depth estimation using adaptive bins.
\newblock In \emph{Proceedings of the IEEE/CVF Conference on Computer Vision
  and Pattern Recognition}, pages 4009--4018, 2021.

\bibitem[Bian et~al.(2019)Bian, Li, Wang, Zhan, Shen, Cheng, and
  Reid]{bian2019unsupervised}
Jiawang Bian, Zhichao Li, Naiyan Wang, Huangying Zhan, Chunhua Shen, Ming-Ming
  Cheng, and Ian Reid.
\newblock Unsupervised scale-consistent depth and ego-motion learning from
  monocular video.
\newblock \emph{Advances in neural information processing systems}, 32, 2019.

\bibitem[Cabon et~al.(2020)Cabon, Murray, and Humenberger]{cabon2020vkitti2}
Yohann Cabon, Naila Murray, and Martin Humenberger.
\newblock Virtual kitti 2, 2020.

\bibitem[Caesar et~al.(2020)Caesar, Bankiti, Lang, Vora, Liong, Xu, Krishnan,
  Pan, Baldan, and Beijbom]{caesar2020nuscenes}
Holger Caesar, Varun Bankiti, Alex~H Lang, Sourabh Vora, Venice~Erin Liong,
  Qiang Xu, Anush Krishnan, Yu Pan, Giancarlo Baldan, and Oscar Beijbom.
\newblock nuscenes: A multimodal dataset for autonomous driving.
\newblock In \emph{Proceedings of the IEEE/CVF conference on computer vision
  and pattern recognition}, pages 11621--11631, 2020.

\bibitem[Chawla et~al.(2021)Chawla, Varma, Arani, and
  Zonooz]{chawla2021multimodal}
Hemang Chawla, Arnav Varma, Elahe Arani, and Bahram Zonooz.
\newblock Multimodal scale consistency and awareness for monocular
  self-supervised depth estimation.
\newblock In \emph{2021 IEEE International Conference on Robotics and
  Automation (ICRA)}, pages 5140--5146. IEEE, 2021.

\bibitem[Chen et~al.(2023)Chen, Li, Zhang, and Li]{chen2023frequency}
Xingyu Chen, Thomas~H Li, Ruonan Zhang, and Ge Li.
\newblock Frequency-aware self-supervised monocular depth estimation.
\newblock In \emph{Proceedings of the IEEE/CVF Winter Conference on
  Applications of Computer Vision}, pages 5808--5817, 2023.

\bibitem[Cho et~al.(2021)Cho, Min, Kim, and Sohn]{cho2021deep}
Jaehoon Cho, Dongbo Min, Youngjung Kim, and Kwanghoon Sohn.
\newblock Deep monocular depth estimation leveraging a large-scale outdoor
  stereo dataset.
\newblock \emph{Expert Systems with Applications}, 178:\penalty0 114877, 2021.

\bibitem[Dang et~al.(2008)Dang, Peng, Wang, and Zhang]{dang2008theil}
Xin Dang, Hanxiang Peng, Xueqin Wang, and Heping Zhang.
\newblock Theil-sen estimators in a multiple linear regression model.
\newblock \emph{Olemiss Edu}, 2008.

\bibitem[Diaz et~al.(2017)Diaz, Walker, Szafir, and Szafir]{diaz2017designing}
Catherine Diaz, Michael Walker, Danielle~Albers Szafir, and Daniel Szafir.
\newblock Designing for depth perceptions in augmented reality.
\newblock In \emph{2017 IEEE international symposium on mixed and augmented
  reality (ISMAR)}, pages 111--122. IEEE, 2017.

\bibitem[Dong et~al.(2022)Dong, Garratt, Anavatti, and Abbass]{dong2022towards}
Xingshuai Dong, Matthew~A Garratt, Sreenatha~G Anavatti, and Hussein~A Abbass.
\newblock Towards real-time monocular depth estimation for robotics: A survey.
\newblock \emph{IEEE Transactions on Intelligent Transportation Systems},
  23\penalty0 (10):\penalty0 16940--16961, 2022.

\bibitem[Eigen et~al.(2014)Eigen, Puhrsch, and Fergus]{eigen2014depth}
David Eigen, Christian Puhrsch, and Rob Fergus.
\newblock Depth map prediction from a single image using a multi-scale deep
  network.
\newblock \emph{Advances in neural information processing systems}, 27, 2014.

\bibitem[Faria and Moreira(2021)]{faria2021implementation}
Jo{\~a}o~M Faria and Ant{\'o}nio~HJ Moreira.
\newblock Implementation of an autonomous ros-based mobile robot with ai depth
  estimation.
\newblock In \emph{IECON 2021--47th Annual Conference of the IEEE Industrial
  Electronics Society}, pages 1--6. IEEE, 2021.

\bibitem[Fu et~al.(2018)Fu, Gong, Wang, Batmanghelich, and Tao]{fu2018deep}
Huan Fu, Mingming Gong, Chaohui Wang, Kayhan Batmanghelich, and Dacheng Tao.
\newblock Deep ordinal regression network for monocular depth estimation.
\newblock In \emph{Proceedings of the IEEE conference on computer vision and
  pattern recognition}, pages 2002--2011, 2018.

\bibitem[Gaidon et~al.(2016)Gaidon, Wang, Cabon, and
  Vig]{Gaidon:Virtual:CVPR2016}
A Gaidon, Q Wang, Y Cabon, and E Vig.
\newblock Virtual worlds as proxy for multi-object tracking analysis.
\newblock In \emph{CVPR}, 2016.

\bibitem[Garg et~al.(2016)Garg, Bg, Carneiro, and Reid]{garg2016unsupervised}
Ravi Garg, Vijay~Kumar Bg, Gustavo Carneiro, and Ian Reid.
\newblock Unsupervised cnn for single view depth estimation: Geometry to the
  rescue.
\newblock In \emph{Computer Vision--ECCV 2016: 14th European Conference,
  Amsterdam, The Netherlands, October 11-14, 2016, Proceedings, Part VIII 14},
  pages 740--756. Springer, 2016.

\bibitem[Geiger et~al.(2013)Geiger, Lenz, Stiller, and
  Urtasun]{geiger2013vision}
Andreas Geiger, Philip Lenz, Christoph Stiller, and Raquel Urtasun.
\newblock Vision meets robotics: The kitti dataset.
\newblock \emph{The International Journal of Robotics Research}, 32\penalty0
  (11):\penalty0 1231--1237, 2013.

\bibitem[Godard et~al.(2019)Godard, Mac~Aodha, Firman, and
  Brostow]{godard2019digging}
Cl{\'e}ment Godard, Oisin Mac~Aodha, Michael Firman, and Gabriel~J Brostow.
\newblock Digging into self-supervised monocular depth estimation.
\newblock In \emph{Proceedings of the IEEE/CVF International Conference on
  Computer Vision}, pages 3828--3838, 2019.

\bibitem[Gordon et~al.(2019)Gordon, Li, Jonschkowski, and
  Angelova]{gordon2019depth}
Ariel Gordon, Hanhan Li, Rico Jonschkowski, and Anelia Angelova.
\newblock Depth from videos in the wild: Unsupervised monocular depth learning
  from unknown cameras.
\newblock In \emph{Proceedings of the IEEE/CVF International Conference on
  Computer Vision}, pages 8977--8986, 2019.

\bibitem[Guizilini et~al.(2020{\natexlab{a}})Guizilini, Ambrus, Pillai,
  Raventos, and Gaidon]{guizilini20203d}
Vitor Guizilini, Rares Ambrus, Sudeep Pillai, Allan Raventos, and Adrien
  Gaidon.
\newblock 3d packing for self-supervised monocular depth estimation.
\newblock In \emph{Proceedings of the IEEE/CVF conference on computer vision
  and pattern recognition}, pages 2485--2494, 2020{\natexlab{a}}.

\bibitem[Guizilini et~al.(2020{\natexlab{b}})Guizilini, Li, Ambrus, Pillai, and
  Gaidon]{guizilini2020robust}
Vitor Guizilini, Jie Li, Rares Ambrus, Sudeep Pillai, and Adrien Gaidon.
\newblock Robust semi-supervised monocular depth estimation with reprojected
  distances.
\newblock In \emph{Conference on robot learning}, pages 503--512. PMLR,
  2020{\natexlab{b}}.

\bibitem[Guizilini et~al.(2021{\natexlab{a}})Guizilini, Ambrus, Burgard, and
  Gaidon]{guizilini2021sparse}
Vitor Guizilini, Rares Ambrus, Wolfram Burgard, and Adrien Gaidon.
\newblock Sparse auxiliary networks for unified monocular depth prediction and
  completion.
\newblock In \emph{Proceedings of the IEEE/CVF Conference on Computer Vision
  and Pattern Recognition}, pages 11078--11088, 2021{\natexlab{a}}.

\bibitem[Guizilini et~al.(2021{\natexlab{b}})Guizilini, Li, Ambruș, and
  Gaidon]{guizilini2021geometric}
Vitor Guizilini, Jie Li, Rareș Ambruș, and Adrien Gaidon.
\newblock Geometric unsupervised domain adaptation for semantic segmentation.
\newblock In \emph{Proceedings of the IEEE/CVF International Conference on
  Computer Vision}, pages 8537--8547, 2021{\natexlab{b}}.

\bibitem[Guizilini et~al.(2022{\natexlab{a}})Guizilini, Lee, Ambru{\c{s}}, and
  Gaidon]{guizilini2022learning}
Vitor Guizilini, Kuan-Hui Lee, Rare{\c{s}} Ambru{\c{s}}, and Adrien Gaidon.
\newblock Learning optical flow, depth, and scene flow without real-world
  labels.
\newblock \emph{IEEE Robotics and Automation Letters}, 7\penalty0 (2):\penalty0
  3491--3498, 2022{\natexlab{a}}.

\bibitem[Guizilini et~al.(2022{\natexlab{b}})Guizilini, Vasiljevic, Ambrus,
  Shakhnarovich, and Gaidon]{guizilini2022full}
Vitor Guizilini, Igor Vasiljevic, Rares Ambrus, Greg Shakhnarovich, and Adrien
  Gaidon.
\newblock Full surround monodepth from multiple cameras.
\newblock \emph{IEEE Robotics and Automation Letters}, 7\penalty0 (2):\penalty0
  5397--5404, 2022{\natexlab{b}}.

\bibitem[Guizilini et~al.(2023)Guizilini, Vasiljevic, Chen, Ambruș, and
  Gaidon]{guizilini2023towards}
Vitor Guizilini, Igor Vasiljevic, Dian Chen, Rareș Ambruș, and Adrien Gaidon.
\newblock Towards zero-shot scale-aware monocular depth estimation.
\newblock In \emph{Proceedings of the IEEE/CVF International Conference on
  Computer Vision}, pages 9233--9243, 2023.

\bibitem[Guo et~al.(2018)Guo, Li, Yi, Ren, and Wang]{guo2018learning}
Xiaoyang Guo, Hongsheng Li, Shuai Yi, Jimmy Ren, and Xiaogang Wang.
\newblock Learning monocular depth by distilling cross-domain stereo networks.
\newblock In \emph{Proceedings of the European Conference on Computer Vision
  (ECCV)}, pages 484--500, 2018.

\bibitem[Hartley and Zisserman(2003)]{hartley2003multiple}
Richard Hartley and Andrew Zisserman.
\newblock \emph{Multiple view geometry in computer vision}.
\newblock Cambridge university press, 2003.

\bibitem[Kanbara et~al.(2000)Kanbara, Okuma, Takemura, and
  Yokoya]{kanbara2000stereoscopic}
Masayuki Kanbara, Takashi Okuma, Haruo Takemura, and Naokazu Yokoya.
\newblock A stereoscopic video see-through augmented reality system based on
  real-time vision-based registration.
\newblock In \emph{Proceedings IEEE Virtual Reality 2000 (Cat. No. 00CB37048)},
  pages 255--262. IEEE, 2000.

\bibitem[Kendall et~al.(2017)Kendall, Martirosyan, Dasgupta, Henry, Kennedy,
  Bachrach, and Bry]{kendall2017end}
Alex Kendall, Hayk Martirosyan, Saumitro Dasgupta, Peter Henry, Ryan Kennedy,
  Abraham Bachrach, and Adam Bry.
\newblock End-to-end learning of geometry and context for deep stereo
  regression.
\newblock In \emph{Proceedings of the IEEE international conference on computer
  vision}, pages 66--75, 2017.

\bibitem[Kim et~al.(2021)Kim, Lee, Kim, Kim, Lee, and Choi]{kim2021stereo}
Wan-Soo Kim, Dae-Hyun Lee, Yong-Joo Kim, Taehyeong Kim, Won-Suk Lee, and
  Chang-Hyun Choi.
\newblock Stereo-vision-based crop height estimation for agricultural robots.
\newblock \emph{Computers and Electronics in Agriculture}, 181:\penalty0
  105937, 2021.

\bibitem[Kuznietsov et~al.(2017)Kuznietsov, Stuckler, and
  Leibe]{kuznietsov2017semi}
Yevhen Kuznietsov, Jorg Stuckler, and Bastian Leibe.
\newblock Semi-supervised deep learning for monocular depth map prediction.
\newblock In \emph{Proceedings of the IEEE conference on computer vision and
  pattern recognition}, pages 6647--6655, 2017.

\bibitem[Lee et~al.(2019)Lee, Han, Ko, and Suh]{lee2019big}
Jin~Han Lee, Myung-Kyu Han, Dong~Wook Ko, and Il~Hong Suh.
\newblock From big to small: Multi-scale local planar guidance for monocular
  depth estimation.
\newblock \emph{arXiv preprint arXiv:1907.10326}, 2019.

\bibitem[Lo et~al.(2022)Lo, Wang, Thomas, Zheng, Patel, and
  Kuo]{lo2022learning}
Shao-Yuan Lo, Wei Wang, Jim Thomas, Jingjing Zheng, Vishal~M Patel, and
  Cheng-Hao Kuo.
\newblock Learning feature decomposition for domain adaptive monocular depth
  estimation.
\newblock \emph{arXiv preprint arXiv:2208.00160}, 2022.

\bibitem[McCraith et~al.(2020)McCraith, Neumann, and
  Vedaldi]{mccraith2020calibrating}
Robert McCraith, Lukas Neumann, and Andrea Vedaldi.
\newblock Calibrating self-supervised monocular depth estimation.
\newblock \emph{arXiv preprint arXiv:2009.07714}, 2020.

\bibitem[Nidamanuri et~al.(2021)Nidamanuri, Nibhanupudi, Assfalg, and
  Venkataraman]{nidamanuri2021progressive}
Jaswanth Nidamanuri, Chinmayi Nibhanupudi, Rolf Assfalg, and Hrishikesh
  Venkataraman.
\newblock A progressive review: Emerging technologies for adas driven
  solutions.
\newblock \emph{IEEE Transactions on Intelligent Vehicles}, 7\penalty0
  (2):\penalty0 326--341, 2021.

\bibitem[PD(2021)]{PD}
PD.
\newblock Parallel domain.
\newblock \url{https://paralleldomain.com/}, 2021.

\bibitem[Ranftl et~al.(2020)Ranftl, Lasinger, Hafner, Schindler, and
  Koltun]{ranftl2020towards}
Ren{\'e} Ranftl, Katrin Lasinger, David Hafner, Konrad Schindler, and Vladlen
  Koltun.
\newblock Towards robust monocular depth estimation: Mixing datasets for
  zero-shot cross-dataset transfer.
\newblock \emph{IEEE transactions on pattern analysis and machine
  intelligence}, 44\penalty0 (3):\penalty0 1623--1637, 2020.

\bibitem[Shao et~al.(2021)Shao, Pei, Chen, Zhang, Wu, Sun, and
  Doermann]{shao2021self}
Shuwei Shao, Zhongcai Pei, Weihai Chen, Baochang Zhang, Xingming Wu, Dianmin
  Sun, and David Doermann.
\newblock Self-supervised learning for monocular depth estimation on minimally
  invasive surgery scenes.
\newblock In \emph{2021 IEEE International Conference on Robotics and
  Automation (ICRA)}, pages 7159--7165. IEEE, 2021.

\bibitem[Swami et~al.(2022)Swami, Muduli, Gurram, and Bajpai]{swami2022you}
Kunal Swami, Amrit Muduli, Uttam Gurram, and Pankaj Bajpai.
\newblock Do what you can, with what you have: Scale-aware and high quality
  monocular depth estimation without real world labels.
\newblock In \emph{Proceedings of the IEEE/CVF Conference on Computer Vision
  and Pattern Recognition}, pages 988--997, 2022.

\bibitem[Uhrig et~al.(2017)Uhrig, Schneider, Schneider, Franke, Brox, and
  Geiger]{uhrig2017sparsity}
Jonas Uhrig, Nick Schneider, Lukas Schneider, Uwe Franke, Thomas Brox, and
  Andreas Geiger.
\newblock Sparsity invariant cnns.
\newblock In \emph{2017 international conference on 3D Vision (3DV)}, pages
  11--20. IEEE, 2017.

\bibitem[Wang et~al.(2018)Wang, Buenaposada, Zhu, and Lucey]{wang2018learning}
Chaoyang Wang, Jos{\'e}~Miguel Buenaposada, Rui Zhu, and Simon Lucey.
\newblock Learning depth from monocular videos using direct methods.
\newblock In \emph{Proceedings of the IEEE conference on computer vision and
  pattern recognition}, pages 2022--2030, 2018.

\bibitem[Watson et~al.(2019)Watson, Firman, Brostow, and
  Turmukhambetov]{watson2019self}
Jamie Watson, Michael Firman, Gabriel~J Brostow, and Daniyar Turmukhambetov.
\newblock Self-supervised monocular depth hints.
\newblock In \emph{Proceedings of the IEEE/CVF International Conference on
  Computer Vision}, pages 2162--2171, 2019.

\bibitem[Watson et~al.(2021)Watson, Mac~Aodha, Prisacariu, Brostow, and
  Firman]{watson2021temporal}
Jamie Watson, Oisin Mac~Aodha, Victor Prisacariu, Gabriel Brostow, and Michael
  Firman.
\newblock The temporal opportunist: Self-supervised multi-frame monocular
  depth.
\newblock In \emph{Proceedings of the IEEE/CVF Conference on Computer Vision
  and Pattern Recognition}, pages 1164--1174, 2021.

\bibitem[Wei et~al.(2023)Wei, Zhao, Zheng, Zhu, Rao, Huang, Lu, and
  Zhou]{wei2023surrounddepth}
Yi Wei, Linqing Zhao, Wenzhao Zheng, Zheng Zhu, Yongming Rao, Guan Huang, Jiwen
  Lu, and Jie Zhou.
\newblock Surrounddepth: Entangling surrounding views for self-supervised
  multi-camera depth estimation.
\newblock In \emph{Conference on Robot Learning}, pages 539--549. PMLR, 2023.

\bibitem[Xiao et~al.(2020)Xiao, Codevilla, Gurram, Urfalioglu, and
  L{\'o}pez]{xiao2020multimodal}
Yi Xiao, Felipe Codevilla, Akhil Gurram, Onay Urfalioglu, and Antonio~M
  L{\'o}pez.
\newblock Multimodal end-to-end autonomous driving.
\newblock \emph{IEEE Transactions on Intelligent Transportation Systems},
  23\penalty0 (1):\penalty0 537--547, 2020.

\bibitem[Xue et~al.(2020)Xue, Zhuo, Huang, Fu, Wu, and Ang]{xue2020toward}
Feng Xue, Guirong Zhuo, Ziyuan Huang, Wufei Fu, Zhuoyue Wu, and Marcelo~H Ang.
\newblock Toward hierarchical self-supervised monocular absolute depth
  estimation for autonomous driving applications.
\newblock In \emph{2020 IEEE/RSJ International Conference on Intelligent Robots
  and Systems (IROS)}, pages 2330--2337. IEEE, 2020.

\bibitem[Zhang et~al.(2022)Zhang, Zhang, and Tao]{zhang2022towards}
Sen Zhang, Jing Zhang, and Dacheng Tao.
\newblock Towards scale-aware, robust, and generalizable unsupervised monocular
  depth estimation by integrating imu motion dynamics.
\newblock In \emph{European Conference on Computer Vision}, pages 143--160.
  Springer, 2022.

\bibitem[Zhao et~al.(2019)Zhao, Fu, Gong, and Tao]{zhao2019geometry}
Shanshan Zhao, Huan Fu, Mingming Gong, and Dacheng Tao.
\newblock Geometry-aware symmetric domain adaptation for monocular depth
  estimation.
\newblock In \emph{Proceedings of the IEEE/CVF Conference on Computer Vision
  and Pattern Recognition}, pages 9788--9798, 2019.

\bibitem[Zheng et~al.(2018)Zheng, Cham, and Cai]{zheng2018t2net}
Chuanxia Zheng, Tat-Jen Cham, and Jianfei Cai.
\newblock T2net: Synthetic-to-realistic translation for solving single-image
  depth estimation tasks.
\newblock In \emph{Proceedings of the European conference on computer vision
  (ECCV)}, pages 767--783, 2018.

\bibitem[Zhou et~al.(2017)Zhou, Brown, Snavely, and Lowe]{zhou2017unsupervised}
Tinghui Zhou, Matthew Brown, Noah Snavely, and David~G Lowe.
\newblock Unsupervised learning of depth and ego-motion from video.
\newblock In \emph{Proceedings of the IEEE conference on computer vision and
  pattern recognition}, pages 1851--1858, 2017.

\end{thebibliography}


\begin{thebibliography}{21}
\providecommand{\natexlab}[1]{#1}
\providecommand{\url}[1]{\texttt{#1}}
\expandafter\ifx\csname urlstyle\endcsname\relax
  \providecommand{\doi}[1]{doi: #1}\else
  \providecommand{\doi}{doi: \begingroup \urlstyle{rm}\Url}\fi

\bibitem[Cabon et~al.(2020)Cabon, Murray, and Humenberger]{cabon2020vkitti2}
Yohann Cabon, Naila Murray, and Martin Humenberger.
\newblock Virtual kitti 2, 2020.

\bibitem[Caesar et~al.(2020)Caesar, Bankiti, Lang, Vora, Liong, Xu, Krishnan,
  Pan, Baldan, and Beijbom]{caesar2020nuscenes}
Holger Caesar, Varun Bankiti, Alex~H Lang, Sourabh Vora, Venice~Erin Liong,
  Qiang Xu, Anush Krishnan, Yu Pan, Giancarlo Baldan, and Oscar Beijbom.
\newblock nuscenes: A multimodal dataset for autonomous driving.
\newblock In \emph{Proceedings of the IEEE/CVF conference on computer vision
  and pattern recognition}, pages 11621--11631, 2020.

\bibitem[Casser et~al.(2019)Casser, Pirk, Mahjourian, and
  Angelova]{casser2019depth}
Vincent Casser, Soeren Pirk, Reza Mahjourian, and Anelia Angelova.
\newblock Depth prediction without the sensors: Leveraging structure for
  unsupervised learning from monocular videos.
\newblock In \emph{Proceedings of the AAAI conference on artificial
  intelligence}, pages 8001--8008, 2019.

\bibitem[Dang et~al.(2008)Dang, Peng, Wang, and Zhang]{dang2008theil}
Xin Dang, Hanxiang Peng, Xueqin Wang, and Heping Zhang.
\newblock Theil-sen estimators in a multiple linear regression model.
\newblock \emph{Olemiss Edu}, 2008.

\bibitem[Eigen et~al.(2014)Eigen, Puhrsch, and Fergus]{eigen2014depth}
David Eigen, Christian Puhrsch, and Rob Fergus.
\newblock Depth map prediction from a single image using a multi-scale deep
  network.
\newblock \emph{Advances in neural information processing systems}, 27, 2014.

\bibitem[Gaidon et~al.(2016)Gaidon, Wang, Cabon, and
  Vig]{Gaidon:Virtual:CVPR2016}
A Gaidon, Q Wang, Y Cabon, and E Vig.
\newblock Virtual worlds as proxy for multi-object tracking analysis.
\newblock In \emph{CVPR}, 2016.

\bibitem[Geiger et~al.(2013)Geiger, Lenz, Stiller, and
  Urtasun]{geiger2013vision}
Andreas Geiger, Philip Lenz, Christoph Stiller, and Raquel Urtasun.
\newblock Vision meets robotics: The kitti dataset.
\newblock \emph{The International Journal of Robotics Research}, 32\penalty0
  (11):\penalty0 1231--1237, 2013.

\bibitem[Godard et~al.(2019)Godard, Mac~Aodha, Firman, and
  Brostow]{godard2019digging}
Cl{\'e}ment Godard, Oisin Mac~Aodha, Michael Firman, and Gabriel~J Brostow.
\newblock Digging into self-supervised monocular depth estimation.
\newblock In \emph{Proceedings of the IEEE/CVF international conference on
  computer vision}, pages 3828--3838, 2019.

\bibitem[Guizilini et~al.(2020)Guizilini, Ambrus, Pillai, Raventos, and
  Gaidon]{guizilini20203d}
Vitor Guizilini, Rares Ambrus, Sudeep Pillai, Allan Raventos, and Adrien
  Gaidon.
\newblock 3d packing for self-supervised monocular depth estimation.
\newblock In \emph{Proceedings of the IEEE/CVF conference on computer vision
  and pattern recognition}, pages 2485--2494, 2020.

\bibitem[Guizilini et~al.(2022)Guizilini, Lee, Ambru{\c{s}}, and
  Gaidon]{guizilini2022learning}
Vitor Guizilini, Kuan-Hui Lee, Rare{\c{s}} Ambru{\c{s}}, and Adrien Gaidon.
\newblock Learning optical flow, depth, and scene flow without real-world
  labels.
\newblock \emph{IEEE Robotics and Automation Letters}, 7\penalty0 (2):\penalty0
  3491--3498, 2022.

\bibitem[Hartley and Zisserman(2003)]{hartley2003multiple}
Richard Hartley and Andrew Zisserman.
\newblock \emph{Multiple view geometry in computer vision}.
\newblock Cambridge university press, 2003.

\bibitem[He et~al.(2016)He, Zhang, Ren, and Sun]{he2016deep}
Kaiming He, Xiangyu Zhang, Shaoqing Ren, and Jian Sun.
\newblock Deep residual learning for image recognition.
\newblock In \emph{Proceedings of the IEEE conference on computer vision and
  pattern recognition}, pages 770--778, 2016.

\bibitem[Kendall et~al.(2017)Kendall, Martirosyan, Dasgupta, Henry, Kennedy,
  Bachrach, and Bry]{kendall2017end}
Alex Kendall, Hayk Martirosyan, Saumitro Dasgupta, Peter Henry, Ryan Kennedy,
  Abraham Bachrach, and Adam Bry.
\newblock End-to-end learning of geometry and context for deep stereo
  regression.
\newblock In \emph{Proceedings of the IEEE international conference on computer
  vision}, pages 66--75, 2017.

\bibitem[Li et~al.(2021)Li, Gordon, Zhao, Casser, and
  Angelova]{li2021unsupervised}
Hanhan Li, Ariel Gordon, Hang Zhao, Vincent Casser, and Anelia Angelova.
\newblock Unsupervised monocular depth learning in dynamic scenes.
\newblock In \emph{Conference on Robot Learning}, pages 1908--1917. PMLR, 2021.

\bibitem[Sandler et~al.(2018)Sandler, Howard, Zhu, Zhmoginov, and
  Chen]{sandler2018mobilenetv2}
Mark Sandler, Andrew Howard, Menglong Zhu, Andrey Zhmoginov, and Liang-Chieh
  Chen.
\newblock Mobilenetv2: Inverted residuals and linear bottlenecks.
\newblock In \emph{Proceedings of the IEEE conference on computer vision and
  pattern recognition}, pages 4510--4520, 2018.

\bibitem[Swami et~al.(2022)Swami, Muduli, Gurram, and Bajpai]{swami2022you}
Kunal Swami, Amrit Muduli, Uttam Gurram, and Pankaj Bajpai.
\newblock Do what you can, with what you have: Scale-aware and high quality
  monocular depth estimation without real world labels.
\newblock In \emph{Proceedings of the IEEE/CVF Conference on Computer Vision
  and Pattern Recognition}, pages 988--997, 2022.

\bibitem[Tao et~al.(2020)Tao, Sapra, and Catanzaro]{tao2020hierarchical}
Andrew Tao, Karan Sapra, and Bryan Catanzaro.
\newblock Hierarchical multi-scale attention for semantic segmentation.
\newblock \emph{arXiv preprint arXiv:2005.10821}, 2020.

\bibitem[Uhrig et~al.(2017)Uhrig, Schneider, Schneider, Franke, Brox, and
  Geiger]{uhrig2017sparsity}
Jonas Uhrig, Nick Schneider, Lukas Schneider, Uwe Franke, Thomas Brox, and
  Andreas Geiger.
\newblock Sparsity invariant cnns.
\newblock In \emph{2017 international conference on 3D Vision (3DV)}, pages
  11--20. IEEE, 2017.

\bibitem[Watson et~al.(2021)Watson, Mac~Aodha, Prisacariu, Brostow, and
  Firman]{watson2021temporal}
Jamie Watson, Oisin Mac~Aodha, Victor Prisacariu, Gabriel Brostow, and Michael
  Firman.
\newblock The temporal opportunist: Self-supervised multi-frame monocular
  depth.
\newblock In \emph{Proceedings of the IEEE/CVF Conference on Computer Vision
  and Pattern Recognition}, pages 1164--1174, 2021.

\bibitem[Wu et~al.(2019)Wu, Kirillov, Massa, Lo, and
  Girshick]{wu2019detectron2}
Yuxin Wu, Alexander Kirillov, Francisco Massa, Wan-Yen Lo, and Ross Girshick.
\newblock Detectron2.
\newblock \url{https://github.com/facebookresearch/detectron2}, 2019.

\bibitem[Zhou et~al.(2017)Zhou, Brown, Snavely, and Lowe]{zhou2017unsupervised}
Tinghui Zhou, Matthew Brown, Noah Snavely, and David~G Lowe.
\newblock Unsupervised learning of depth and ego-motion from video.
\newblock In \emph{Proceedings of the IEEE conference on computer vision and
  pattern recognition}, pages 1851--1858, 2017.

\end{thebibliography}
